\documentclass{article}
\usepackage{arxiv}

\usepackage[utf8]{inputenc} 
\usepackage[T1]{fontenc}    
\usepackage{url}            
\usepackage{booktabs}       
\usepackage{amsfonts}       
\usepackage{nicefrac}       
\usepackage{microtype}      
\usepackage{lipsum}
\usepackage{cellspace}
\usepackage{alltt}
\usepackage{calc}
\usepackage{mathtools}
\usepackage{booktabs}
\usepackage{url}
\usepackage{caption}
\usepackage{longtable}
\usepackage{amsmath}
\usepackage{amsthm}
\usepackage{amssymb}
\usepackage{bm}
\usepackage{adjustbox}
\usepackage{wrapfig}
\usepackage{epstopdf}
\usepackage{tikz}
\usepackage{framed}
\usepackage{thmtools,thm-restate}
\usepackage{enumitem}
\usepackage{soul}
\usepackage{tcolorbox}
\usepackage{multirow}
\usepackage{array}
\usepackage{makecell}
\usepackage[colorlinks=true,breaklinks]{hyperref}
\usepackage{cleveref}

\usetikzlibrary{arrows}
\usetikzlibrary{shapes,backgrounds}

\usepackage{lipsum,graphicx,multicol}
\usepackage{caption}
\usepackage{subcaption}
\usepackage[figuresright]{rotating}
\usepackage[linesnumbered,ruled,vlined]{algorithm2e}
\graphicspath{{./Images/}}

\DeclareMathOperator*{\argmin}{argmin} 

\newtheorem*{theorem*}{Theorem}

\newtheorem*{lemma*}{Lemma}

\makeatletter
\renewcommand{\th@definition}{%
  \normalfont
  \thm@preskip-2 \relax
  \thm@postskip-2 \relax
}
\makeatother

\title{Selective information exchange in collaborative clustering using regularized Optimal Transport }

\author{Fatima Ezzahraa Ben Bouazza, Youn{\`e}s Bennani \\
   Universit\'e Sorbonne Paris Nord, LIPN UMR 7030 CNRS\\
  LaMSN, La Maison des Sciences Numériques, USPN, France\\
  \url{name.surname@sorbonne-paris-nord.fr}

}

\begin{document}
\maketitle

\begin{abstract}
Collaborative learning has recently achieved very significant results. It still suffers, however, from several issues, including the type of information that needs to be exchanged, the criteria for stopping and how to choose the right collaborators. We aim in this paper to improve the quality of the collaboration and to resolve these issues via a novel approach inspired by Optimal Transport theory. More specifically, the objective function for the exchange of information is based on the  Wasserstein distance, with a bidirectional transport of information between collaborators. This formulation allows to learns a stopping criterion and provide a criterion to choose the best collaborators. Extensive experiments are conducted on multiple data-sets to evaluate the proposed approach.
\end{abstract}

\maketitle

\section{Introduction}
\label{intro}

Data clustering is one of the main interests in unsupervised Machine Learning research \cite{kotsiantis2004recent}. A large number of clustering algorithms have been proposed in the literature \cite{wu2009adapting}, divided into different families based on the cost function to optimize \cite{yarowsky1995unsupervised, kotsiantis2004recent}. 

Clustering task is known to be difficult and suffer from several issues. Most of the problems come from the fact that unsupervised algorithms work with very little information about the expected result \cite{yang2006unsupervised}. Therefore, the choice of the cost function to optimize, the algorithm to use and the values of the parameters require a lot of expertise to obtain the desired output \cite{dougherty1995supervised}. In addition, modern data-sets are often very large (both in size and dimension) and distributed into several sites \cite{yang2018multi}, which limit the efficiency of most classical clustering algorithms \cite{grira2004unsupervised}. 

In an attempt to solve these issues, the scientific community has suggested several ways of combining the results of different algorithms \cite{xu2013survey}. Several approaches have been proposed in that direction, based on the idea of several algorithms working on the data, either with each algorithm optimizing a different cost function or working with different values of the parameters on the same data-set, or with each algorithm working on a subset of the data, usually trying to optimize the same cost functions. These approaches can be classified into two main categories: In Ensemble Learning approaches, several algorithms are trained on the data and the set of results are merged into a global consensus \cite{wu2014k}. In Collaborative Clustering several models are trained simultaneously on the data-set, usually each algorithm working on a sub-set of the data, and exchange information during the learning process \cite{forestier2007collaborative}. In this paper we focus on the later approach.

Generally speaking, the problem of Collaborative Clustering can be defined as follows: Given a finite number of disjoint data sites, collaborative clustering is a scheme of collective development and reconciliation of fundamental cluster structures across these sites \cite{pedrycz2002collaborative}.
The general framework for collaborative clustering is based on two principal steps: 

\bigskip
\noindent\textbf{Local step}: Each algorithm will train on the data it has access to and produce a clustering result, e.g. a model of the local data subset. 

\bigskip
\noindent\textbf{Collaborative step}: The algorithms share their output in order to confirm or improve their models, with the goal of finding better clustering solutions.

\bigskip

In this paper, we propose to study the unsupervised collaboration framework through the  Optimal Transport theory, thus benefiting from this mathematical formalism to analyze and describe the process of collaboration between the different algorithms. In this case, the collaboration, that consists of exchanges of information between algorithms, will be modeled in the form of bi-directional or even multi-directional transports. 

The rest of the paper is organized as follow. Section \ref{related} includes the prototype based methods proposed in collaborative learning task, section \ref{sec:fundamental-background } develops the background of  Optimal Transport theory. In Section \ref{sec:Proposed} we introduce the novel framework of collaborative clustering using  Optimal Transport theory. In Section \ref{sec:Experiment} we provide an experimental validation and discuss the quality of the proposed approach. Finally, in Section \ref{sec:Conclusion} a conclusion and some perspective work are given.

\section{Related work}
\label{related}

The first Collaborative clustering was introduced by \cite{pedrycz2002collaborative} under the name "Collaborative Fuzzy Clustering” (CoFC). This approach was based on extended version of Fuzzy C-means adapted to distributed data. The algorithm was based on two steps, the first step aims to find $c$ clusters for each collaborator where each object is assigned to some cluster with a certain degree membership stored in matrix $S$. The second step consists to exchange the information stored in the matrix $S$ or the prototypes of each cluster. The algorithm of Fuzzy C-Means is trained again for each collaborator taking into account the shared information.

Several studies had been done to develop several algorithms and approaches on this framework, such as CoEM in \cite{bickel2005estimation}, CoFKM \cite{cleuziou2009cofkm} and collaborative EM-like algorithm (EM for Expectation–Maximization) based on Markov Random Fields \cite{hu2006maximum}. All these approaches follow the same principle as Collaborative Fuzzy C-Means.

However, these algorithms display similar limitations: they   require the same number of clusters in each site, the same same model trained in each site, and the algorithm can only happen between instances of the same algorithm.  

The collaborative clustering was also developed based on Self-Organization Maps (SOM) \cite{grozavu2010topological} by adapting the original objective function to distributed data. The main idea was to add a term inspired by the classical SOM neighborhood function to the original SOM objective function, where this term aims to compare neighborhoods of each prototype in each sites. This neighborhood term is adaptable to either horizontal or vertical collaboration. The same principle can also be adapted to the Generative Topographic Maps (GTM) \cite{ghassany2012collaborative} with a modification in the M-step of the EM algorithm. The modification consists to add collaborative term inspired from the penalized likelihood estimation \cite{green1990use}.

Another approach was proposed in this framework is the SAMARAH algorithm \cite{wemmert2000classification, forestier2007collaborative}, with the advantage of not requiring a smoothness function or the same number of clusters or prototypes.  However, it is restricted to horizontal collaboration only and the principle of solving the conflict  based on pairwise criterion can make the process volatile.

Recent works have been done to develop the collaborative clustering and make it more flexible \cite{lachaud2017collaborative, sublime2017entropy}, ensuring a collaboration between different algorithms without fixing a unique number of clusters for all of the collaborators. The advantage of this approach is that different families of clustering algorithms can exchange information in a collaborative framework. Nevertheless, one of the most important issue in collaborative clustering is the control on the quality on the exchanged information from several collaborators and the right time to stop the collaboration. In \cite{rastin2015collaborative,bouazza2020collaborative}, the authors develop a new criterion to select the optimal collaborator. They showed that the diversity between collaborators could be an important impact on collaboration. Furthermore, recently a study of the influence of diversity on the collaboration was done based on the entropy \cite{sublime2019study} and showed trade-off between the gain quality and diversity between the collaborators.

\section{Fundamental background of the proposed approach}
\label{sec:fundamental-background }

In this section we will represents the mathematical formalization of  Optimal Transport problem and how it could be resolved using the  Sinkhorn algorithm \cite{cuturi2013sinkhorn}.

\subsection{Optimal Transport}

Optimal Transport is a well defended theory introduced by Monge in \cite{monge1781memoire} to resolve the problem of resources allocation. The basis of this theory was to compute the optimal path of a massive particle from one point to another, by minimizing the cost of this move or this transportation. Lately the Monge problem was relaxed by Kantorovich in \cite{kantorovich2006translocation}, where the problem is transposed to a distribution problem using linear programming connecting a pair of distributions.

More formally, Let $\Omega \subseteq \mathbb{R}^n$ be the measurable space of dimension $n$, and $\mathcal{P}(\Omega)$ denotes the set of probability measures on $\Omega$.

Given two families of data sets in $\Omega$, $X_s=\lbrace x_i^s\rbrace_{i=1}^{n_s}$ and $X_t=\lbrace x_i^t\rbrace_{i=1}^{n_t}$, let $\mu_s$ and $\mu_t$ be their respective distributions over $\Omega_s$ and $\Omega_t$ respectively. 

The transport map $\gamma$ from $\mu_s$ to $\mu_t$ is defined as the pushforward $\gamma_{\#\mu_s}=\mu_t$ :
\begin{equation}
   \gamma: \Omega_s\rightarrow\Omega_t
\end{equation}

where $\gamma$ transforms the probability measure $\mu_s$ in its image measure noted $\gamma_{\#\mu_s}$,  which is another probability measure defined over $\Omega_t$ and satisfying:

\begin{equation}
    \gamma_{\#\mu_s(x)}=\mu_s(\gamma^{-1}(x)), \forall x \in \Omega_t
\end{equation}

The Monge-Kantorovich formulation of this problem is a convex relaxation which aims to find a coupling $\gamma$ defined as a joint probability measure over  $\Omega_s \times\Omega_t$ with marginals $\mu_s$ and $\mu_t$  that minimizes the cost of transport w.r.t $c:\Omega_s\times\Omega_t \rightarrow \mathbb{R_+}$ :

\begin{equation}
   \gamma^*= \argmin_{\gamma\in\Pi} \int_{\Omega_s\times\Omega_t} c(x^s,x^t) d\gamma(x^s,x^t)
\end{equation}

Where $\Pi$ is the set of all probabilistic couplings in $\mathcal{P}(\Omega_s\times\Omega_t)$ with marginals $\mu_s$ and $\mu_t$, and $\gamma^*$ designate the Optimal transportation plan.

This problem admits a unique solution $\gamma^*$ which allows to define the Wasserstein distance of order $p\in [1, +\infty[$ between $\mu_s$ and $\mu_t$ :

\begin{equation}\label{PWS}
W_{p}(\mu_s,\mu_t)=\left( \inf_{\gamma\in\Pi(\mu_s,\mu_t)}\int_{\Omega_s\times\Omega_t} d^p(x^s,x^t)d\gamma(x^s,x^t)\right) ^{\frac{1}{p}}
\end{equation}

where $d$ is a distance corresponding to the cost function $c(x^s,x^t)=d^p(x^s,x^t)$ 

s.t  $c:\Omega_s\times\Omega_t \rightarrow \mathbb{R_+}$  of transporting the unit mass $x^s$ to $x^t$.
\bigskip

In this work, we focus on the discrete case of the  Optimal Transport problem. However, we refer to \cite{villani2008optimal} for more details on the continuous case and the mathematics involved.

\bigskip
We consider the discrete setting of Optimal Transport problem. This case arises where $\mu_s$ and $\mu_t$ are only accessible through discrete samples. The empirical measures can be defined as:

\begin{equation}
    \mu_s=\sum_{i=1}^{n_s}p_i^s\delta_{x_{i}^s}\quad\quad \mathrm{and}\quad\quad \mu_t=\sum_{i=1}^{n_t}p_i^t\delta_{x_{i}^t}
\end{equation}

With\quad$\delta_{x_i}$ the Dirac function at $x_{i}\in\mathbb{R}^n$

\quad\quad\quad$p_i^s$ and $p_i^t$  the probability masses associated to the $ith$ sample, 

\quad\quad\quad s.t $\sum_{i=1}^{n_s}p_i^s=\sum_{i=1}^{n_t}p_i^t=1$.

\bigskip
The Monge-Kantorovich problem consists on finding an optimal coupling (or Transportation plan) $\gamma*$ as a joint probability between $\mu_s$ and $\mu_t$ over $ \Omega_s\times \Omega_t$ by minimizing the cost of the transport w.r.t $X_s\in \mathbb{R}^{n_s\times n}$ and $X_t\in \mathbb{R}^{n_t\times n}$  by solving: 

\begin{equation}
    \gamma^*=\argmin_{\gamma\in\Pi(\mu_s,\mu_t)}< \gamma,C>_F
\end{equation}

with:
\begin{align*}
    \quad\quad&<.,.>_F  ~\text{the Frobenius dot product}\\
    \quad\quad&C\in\mathbb{R}_+^{n_s\times n_t}  ~\text{the transport  cost with $C_{ij}$ given by: $C: X_s\times X_t\rightarrow\mathbb{R}_+$}\\
    \quad\quad&\Pi(\mu_s,\mu_t)=\left\lbrace \gamma \in \mathbb{R}_{+}^{n_s\times n_t} \mid \gamma \mathbf{1}_{n_{s}}=\mu_s, \gamma^{T}\mathbf{1}_{n_{t}}=\mu_t\right\rbrace ~\text{the transportation}\\
    \quad\quad&\text{polytope where $\mathbf{1}_n$ is a n-dimensional vectors of ones}. 
\end{align*}

This problem admits a unique solution $\gamma^{\ast}$ and defines a metric called \textit{the Wasserstein distance} on the space of the discrete  probability measures as follow:

\begin{equation}\label{wsd}
W(\mu_s,\mu_t)=<\gamma^*,C>_F
\end{equation}

The Wasserstein distance has been very useful recently especially in machine learning such as domain adaptation \cite{courty2014domain} metric learning \cite{cuturi2014ground}, clustering \cite{cuturi2014fast} and multi-level clustering \cite{ho2017multilevel,bouazza2019multi}. The particularity about this distance is that it takes into account the geometry of the data using the distance between the samples, which explains its efficiency. On the other hand, in term of computation, the success of this distance also comes from the work of Cuturi \cite{cuturi2013sinkhorn}, who introduced an algorithm based on entropy regularization, as presented in the next section.

\subsection{Regularized Optimal Transport }

Even though the Wasserstein distance has known very significant successes, in term of computation the objective function has always suffered from a very slow convergence, especially in high  dimension, which lead to the idea of proposing a smoothed objective function by adding a term of entropic regularization, introduced in \cite{schwarzschild1916sitzungsberichte} and applied to the Optimal Transport problem in \cite{cuturi2013sinkhorn} in order to speed up the convergence and improve the stability \cite{chizat2020faster}.

This is represented formally by the following minimization problem:
\begin{equation}\label{WR}
\gamma^*_\lambda=\argmin_{\gamma \in \Pi(\mu_s,\mu_t)}<\gamma,C>-\frac{1}{\lambda}E(\gamma)
\end{equation} 
where $E(\gamma)=-\sum_{i,j}^{n_s,n_t}\gamma_{ij}\log(\gamma_{ij})$ and $\lambda >0 $ the entropy regularization parameter and $C$  is the cost matrix.

With the strong convexity of entropy, the objective function became a strictly convex function.   Consequently, the minimization problem (\ref{WR}) admits a unique solution and can be solved by the  Sinkhorn's fixed point algorithm, based on the following theorem.

\bigskip

\textbf{Sinkhorn theorem (1967)}:
For any positive matrix $\mathbf{A} \in \mathcal{M}(\mathbb{R}_{+}^{n\times m)}), a\in \Sigma_{n}$ et $b \in \Sigma_{m}$, there is one and unique pair of vectors $(u,v)\in \mathbb{R}_{+}^{n}\times \mathbb{R}_{+}^{m}$ such that $diag(u)\mathbf{A}diag(v) \in \mathcal{U}(a,b)$ and constitutes a fixed point of the application:
\begin{equation}\label{sink}
 (u,v)\rightarrow(\mathbf{A}v^{-1}./b,\mathbf{A}^{T}u^{-1}./a) 
\end{equation}

Thanks to the regularized version of optimal transport, we obtained a less sparse, smoother and more stable solution than the original problem. Another important advantage is that this formulation allows the scaling matrix approach of Sinkhorn-Knopp \cite{mensch2020online}.

The regularized Optimal Transport plan is then found by iteratively computing two scaling vectors $u$ and $v$ such that 
\begin{equation}\label{sinn}
\gamma^*=diag(u)\exp(\lambda C)diag(v)
\end{equation}

\section{Proposed approach}
\label{sec:Proposed}

\begin{figure}[!h]
  \begin{center}
  \includegraphics[width=\textwidth]{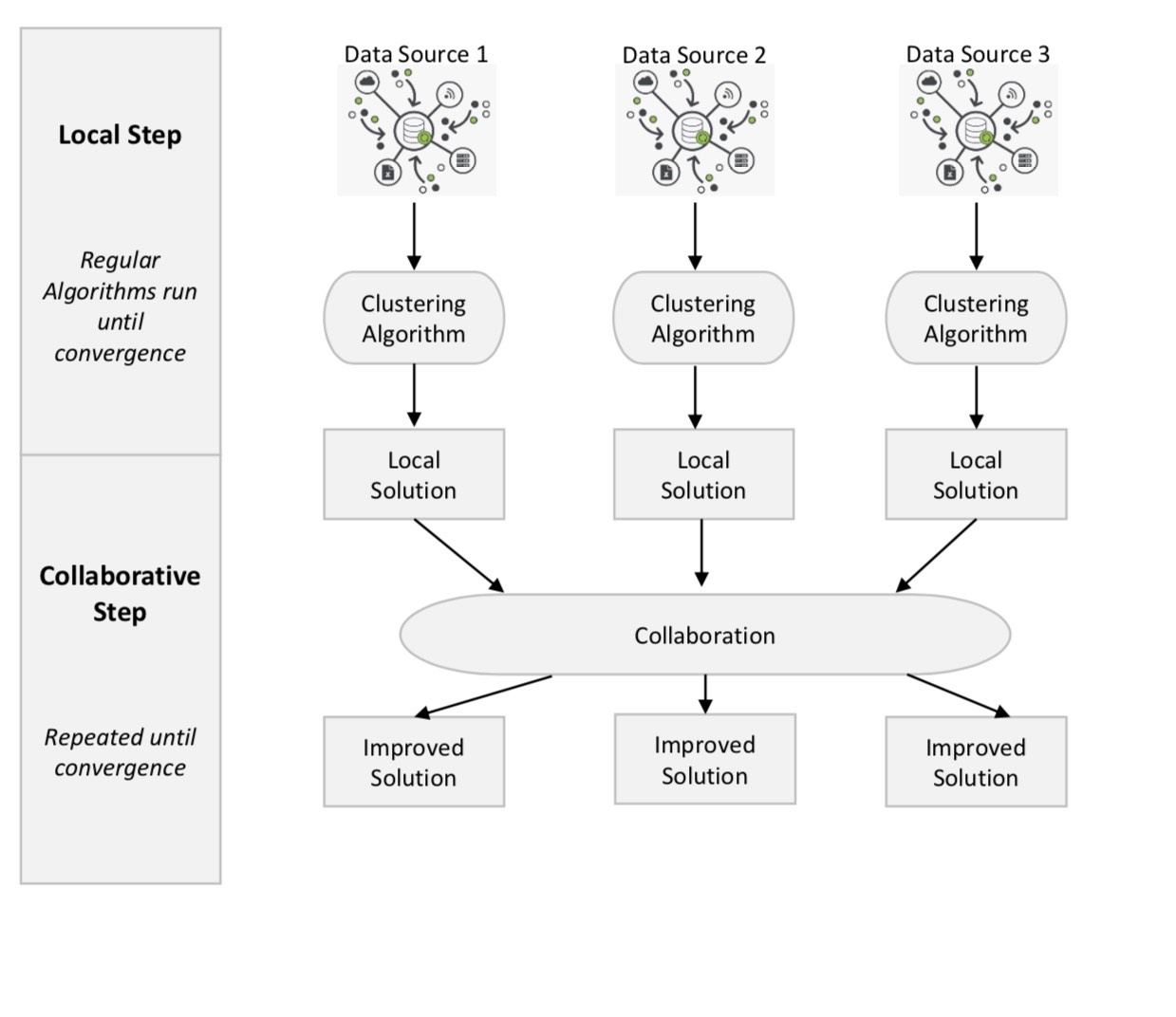}
  \end{center}
  \caption{Collaborative clustering framework with selective information exchange.}
\label{collab2}
\end{figure}

\subsection{Motivation and potential applications}
\label{moti:1}

With the development of hardware technology, a huge amount of data represented in different views and different structures have been generated in real word applications. This kind of data is considered as a new challenge to develop the existing clustering algorithms, designed for single view data, to be more adaptable to multi-view data.

To clarify the  motivation behind the proposed approach, we present some potential industrial applications, where we have several organizations or companies using a collection of data sets that could either concern the same or different customers. This could be data describing customers of banking institutions, state organizations and hospital with medical information records, etc. Imagine that all these organizations are dealing with the same individuals but every organizations may have different characteristics and descriptors for these individuals linked to the activities of the organization. All these organizations may want to explore data mining algorithms on there one data set. On the other hand, they also recognize that as they are other data sets containing information about the same individuals, it would advantageous to learn about the dependencies that they have so that they could reveal a macro-picture. However, due to ethical consideration and privacy issues, these organization are forbidden to share their data sets. which prevents the experts to combine all these data sets into a single view and carrying out different algorithms of classical clustering. For example, the confidentiality requirements in medical records of patients could deny the access into their personal information, and security issues in banking organization forbid to share the customers information. In addition, it may some hesitation from experts about losing the real structure of the data by adding more information and characteristics. In this situation, the exchange of the information through the proposed approach will guarantee the privacy of the information of each organism, and the control of the collaboration to avoid affecting the real structure of the data.

One of the most difficult challenges in collaborative learning is how to choose the right collaborator to collaborate with, which construct the order of the collaboration, not only to increase local quality of each model, but also to ensure the convergence and to avoid over-fitting (Figure  \ref{collab2}).
 
Classical collaborative algorithms are based on two steps. The first one consists to cluster the data locally, the second consists to send and receive information between the local models. Despite the quantity of work on this framework, it still requires many restrictions to ensure the convergence: usually each algorithm must work on the same representation space and must compute the same number of clusters. These restrictions limit the flexibility of collaborative clustering approaches for the analysis of real data.
 
On the other hand, Optimal Transport theory has shown very significant results, especially in transfer learning \cite{courty2014domain} and for comparison of distributions. Based on this idea, our intuition is to model collaborative learning as a bi-directional knowledge transfer and improve the optimization of the cost function based on the comparison of the distributions of local subset, in order to weight the mutual confidence of the collaborators and use a transport plan to transfer the information between them.

In the next section we detail the proposed approach based on Optimal Transport theory, either in vertical or horizontal collaboration.

\subsection{Collaborative Learning algorithms}


The main goal of the proposed approach is to improve the quality and the stability collaboration and guarantee the convergence without over-fitting. In collaborative clustering, we distinct two principal approaches: vertical and horizontal collaboration. In the vertical collaboration, the collaborators learn from different instances represented in the same space, while in horizontal collaboration the collaborators work on the same instances in different representation space. 

In general, different frameworks must be used for vertical and horizontal collaboration. Here we propose a unified framework adapted to both approaches.

\subsubsection{Local step}

Let consider $r$ collaborators, where the data of each collaborator $v$, $X^{v}=\left\lbrace x_{i}^{v}\right\rbrace_{i=1}^{n^{v}}$ with $x_{i}^{v}\in \mathbb{R}^{d^{v}}$ and $d^v$ the dimension of the space representation of $v$, corresponds to a distribution $\mu^{v}=\frac{1}{n^{v}} \sum_{i=1}^{n^v} \delta_{x_{i}^{v}}$.

We seek in the local step to find the centroids $M^{v}=\lbrace m_{1}^{v},..,m_{k^{v}}^{v}\rbrace$, corresponding to a distribution $\nu^{v}$, that represents the local clusters of each collaborator $v$, such as to minimizes the  Optimal Transport plan $L^{v}=\left\lbrace l_{ij}^{v}\right\rbrace_{i,j=1}^{i,j=n^v,k^v}$ between the the local data $X^{v}$ and the centroids  $M^{v}$. 

To achieve this, we will solve the following minimization problem (\ref{WSV}), where the first minimization of $L^{v}$ consists to find the  Optimal Transport plan between the data and the centroids, and the second minimization $M^{v}$ aims to update the distribution of the centroids so that the transport plan is optimal between the data and the centroids. 


\begin{equation}\label{WSV}
\argmin_{L^v\in \Pi(\mu^v,\nu^v), M^v}<L^v,C(X^v,M^{v})>_F-\frac{1}{\lambda}E(L^{v})
\end{equation}

Subject to 
$\sum_{j=1}^{k^{v}}l^{v}_{ij}=\frac{1}{n^{v}}$ and $\sum_{i=1}^{n^{v}}l^{v}_{ij}=\frac{1}{k^{v}} $
and $C: X^v\times M^v\rightarrow\mathbb{R}_+ $ s.t $C_{ij}=c(x^v_i,m^v_i)=\|x_{i}^{v}-m^{v}_{j}\|^{2}$ the euclidean distance between sample $x_i^v$ and the centroids $m_i^v$.

It should be noticed that resolving (\ref{WSV}) is equivalent to a  Lloyd's problem which is the Expectation Minimization  algorithm when $d=1$ and $p=2$ without any constraints on the weights. This is why to resolve this problem we alternate between computing the Sinkhorn matrix $L^{v}$ to assign instances the closest cluster and updating the centroids to decrease the transportation cost.

\begin{algorithm}[!h]

\caption{Sinkhorn-Means local algorithm}
\label{lmv}

\SetKwInOut{Input}{Input}
  \SetKwInOut{Output}{Output}
 \Input{$X^{v}=\left\lbrace x_{i}^{v}\right\rbrace_{i=1}^{n^v}$:  data of collaborator $v$  with distribution $\mu^v$\\
$k_v$ : number of local clusters  \\
$\lambda$ :  entropic constant 
}

\Output{The OT matrix $L^{v}$ and the centroids $M^{v}$}
 
\bigskip 

Initialize the centroids $M^{v}=\left\lbrace m_{j}^{v}\right\rbrace_{j=1}^{k^v}$ randomly \\
Compute the associated distribution $\nu^{v}= \frac{1}{k^{v}}\sum_{j=1}^{k^{v}}\delta_{m_{j}^{v}}$\\

 \Repeat {convergence}{
  Compute the OT matrix $L^v=\left\lbrace l_{ij}^{v}\right\rbrace_{i,j=1}^{i,j=n^v,k^v}$:

  \[(L^{v})^* =  \argmin_{L^{v}\in \Pi(\mu^{v},\nu^{v})} <L^v,C(X^v,M^{v})>_F-\frac{1}{\lambda}E(L^{v})\]   
  
  Update the centroids $M^{v}=\left\lbrace m_{j}^{v}\right\rbrace_{j=1}^{k^v}$:
 \[m^{v}_{j} = \sum_{i}l_{ij}^*x_{i}^{v} \quad 1\leq j\leq k^{v} \]
  
 }
 \Return{$ (L^{v})^* $ and $M^{v}$ }
 
\end{algorithm}

Algorithm \ref{lmv} details the computation of the local objective function (\ref{WSV}), proceeding similarly to $k$-means but with the  advantage of using the Wasserstein distance. This allows to get soft assignment of the data, in contrary to $k$-means, which means that the components of the assignment matrix $l_{ij}\in[0, \frac{1}{n}]$. Besides, the penalty term based on the entropy regularization guarantees a solution with higher entropy which increases the stability of the algorithm and ensures a uniform assignation of the instances.
\subsubsection{Global step}
\label{appro:1}

The global step aims to compute the collaboration between the models where each collaborator can update its local clustering based on information exchanged with the other collaborators, until stabilization of clusters with improved quality. In the proposed approach, the collaboration step could be seen as two simultaneous phases. 

The first phase aims to create an interaction plan based on Sinkhorn matrix distance which compares the local distribution of each collaborator to the others. The idea behind this phase is to allow each model to select the best collaborator to exchange information with, in other words the algorithm will also learns the best order of the collaborations in each iteration. The heuristic work in \cite{rastin2015collaborative} proved that a collaboration with a model proposing a very different data distribution decreases the local quality, while a collaboration between very similar models is ineffective. Thus, the most beneficial collaboration is the one with models of median diversity. Hence, after the construction of the transport plan using the Sinkhorn algorithm, which compares the local structures, the proposed algorithm learns to choose for each model the collaborator with the median distribution similarity.

The second phase consists to exchange information between collaborators to improve local quality of each model. More precisely, we are looking to transport the prototypes to influence the location of the local prototypes; in order to get a higher local quality of each collaborator. 

Considering the same notation above, we seek to minimize the following objective function:



\begin{align}\label{COLW}
\begin{split}
\argmin_{L^{v},M^{v}}\lbrace<L^v,C(X^v,M^{v})>_F- \frac{1}{\lambda}E(L^{v})+&\sum_{v^{'}=1.v'\neq v}^{r}{\alpha_{v',v}}(<L^{v,v'},C(M^{v},M^{v'})>_F \\-&\frac{1}{\lambda}E(L^{v,v'}))\rbrace
\end{split}
\end{align}

Where the first term deals with the local clustering and the remaining is the collaboration term and represents the influence on the local centroids’ distribution by the distant centroid's distributions. $\alpha_{v',v}$ are non-negative coefficients proportional to the  diversity between the collaborators and the difference of local quality, and $L^{v,v'}$ is the  Optimal Transport plan between the centroids of the $v^{th}$ and $v'^{th} $ collaborators.

\begin{algorithm}[!h]
\caption{Collaborative clustering ({Co-OT})}
\label{collot}
  \SetKwInOut{Input}{Input}
  \SetKwInOut{Output}{Output}
  
 \Input{$\left\lbrace X^{v} \right\rbrace_{v=1}^{r}$: the $r$ collaborators' data with distributions $\left\lbrace \mu^{v}\right\rbrace_{v=1}^{r}$ \\
 $\left\lbrace k^{v}\right\rbrace_{v=1}^{r}$  : the numbers of clusters \\
 $\lambda$  : the entropic constant \\
 $\lbrace \alpha_{v,v'}\rbrace_{v,v'=1}^{v,v'=r}$ : the confidence coefficient matrix}
 \Output{The partition matrix $\left\lbrace (L^{v})^* \right\rbrace_{v=1}^{r}$ and  the centroids $ \left\{M^{v}\right\}^{r}_{v=1}$ }

\bigskip
 Initialize the centroids  $M^{v}=\left\lbrace m_{j}^{v}\right\rbrace_{j=1}^{k^v}$ randomly \\
Compute the associated distribution $\nu^{v}= \frac{1}{k^{v}}\sum_{j=1}^{k^{v}}\delta_{m_{j}^{v}}$, $\forall v \in \lbrace 1,...,r\rbrace$\\
 \Repeat{convergence}{
\For{ $v=1,...,r$ }{
  
  Update the centroids $M^{v}$ and the partition matrix $(L^{v})^*$ using a \textbf{local algorithm} (e.g. Sinkhorn-Means, SOM, GTM, $k$-Means...)\\
  
   Update the centroids distribution $\nu^{v}= \frac{1}{k^{v}}\sum_{j=1}^{k^{v}}\delta_{m_{j}^{v}}$ 
  
  \For{$v'=1,...,r$ and $v\neq v'$}{
  Compute the OT matrix $(L^{v,v'})^*=\lbrace l_{jj'}\rbrace_{j,j'=1}^{j,j'=k^{v},k^{v'}}$ between the centroids of collaborators $v$ and $v'$:
    \[(L^{v,v'})^* = \argmin_{L^{v,v'}\in \Pi(\nu^{v},\nu^{v'})}<L^{v,v'},C(M^{v},M^{v'})>_F-\frac{1}{\lambda}E(L^{v,v'})\]   
   } 
   
 Chose the median collaborator: \[v^{*} = {median}_{v'} \left\lbrace (L^{v,v'})^* \right\rbrace_{v'=1}^{r}, v'\neq v\] \\

Update the local centroids based on the collaborator’s information, if the internal quality is increased (see below): \\
 
 
  \[m^{v}_{j} = \alpha_{v,v*}\sum_{j'}l_{jj'}^{v,v*}m_{j'}^{v*} \quad 1\leq j\leq k^{v}\]

}
 

}
 
 \Return{$\left\lbrace (L^{v})^* \right\rbrace_{v=1}^{r}$ and  the centroids $ \left\{M^{v}\right\}^{r}_{v=1}$  }
\end{algorithm}

Algorithm \ref{collot} explains the computation steps of the proposed approach and shows how it learns to select the best collaborator to learn from at each iteration, based on Sinkhorn comparisons between the distributions, and how it alternates between influencing the local centroids based on the confidence coefficient relative to the chosen collaborator and its local centroids distribution and update of the centroids relative to local instances in order to improve the clusters' quality.

It should be pointed out that in each iteration, each collaborator chooses successively the collaborators to exchange information with, based on the Sinkhorn matrix distance. More accurately, in each iteration, each model exchanges information with the collaborator having the median similarity between the two modelled distributions, computed with the Wasserstein metric. If this exchange increases the quality of the model (here we use the Davies-Bouldin index \cite{davies1979cluster}), the centroids of the model are updated. Otherwise, the selected collaborator is removed from the list of possible collaborators and the process is repeated with the remaining collaborators, until the quality of the clusters stops increasing.

It must be highlighted that the proposed algorithm can be adapted to both horizontal a vertical collaboration, since the inputs of the algorithm requires the distributions that represent the local structure of each collaborator, where it can be either  sharing the same space but different samples (vertical collaboration), which formally means that $X^{v}=\left\{x_{i}^{v}\right\}_{i=1}^{n^{v}} $ such that $x_{i}^{v} \in \mathbb{R}^{d}$  or built  from different spaces but share the same instances (horizontal collaboration), which means $X^{v}=\left\{ x_{i}^{v}\right\}_{i=1}^{n}$ such that $x_{i}^{v} \in \mathbb{R}^{d_{v}}$.

Another important advantage of the proposed algorithm is its adaptability with all the prototype-based algorithms, more precisely instead of using the Sinkhorn-Means as a local algorithm to get the centroids, we can use other prototype models like k-means, SOM, EM, etc. The proposed algorithm has therefore the capability to work with hybrid models. This will be detailed in Section \ref{subcompa}.


\section{Experiments}
\label{sec:Experiment}

\subsection{Setting}
 
 \textbf{\subsubsection{Data-sets}}
We consider the following data-sets provided by the UCI Machine Learning Repository \cite{Dua2017}, described in Table \ref{Data}. Each data-set is split between several collaborators. 

\begin{table}[!h]
\caption{Some characteristics of the experimental real-world data-sets}
\label{Data}
\renewcommand{\arraystretch}{1.5}
\setlength{\tabcolsep}{0.7cm}
\centering
\begin{tabular}{ lccc}
\hline
Data-sets \quad \quad & \#instances \quad \quad & \#Attributes \quad \quad& \#Classes \\ \toprule
Glass & 214 & 10 &7 \\
Spambase & 4601 & 57 & 6 \\ 
Waveform-noise & 5000 & 40 & 3 \\ 
Wdbc & 569 & 33 & 2 \\ 
Wine & 178 & 13 & 3 \\ \bottomrule
\end{tabular}
\end{table}

\begin{itemize}
\item Glass dataset represents oxide content of the glass to determine its type. The study of classification of types of glass was motivated by criminological investigation. Since the glass left at the scene of the crime can be used as evidence...if it is correctly identified!
\item The Spam base dataset consists of 57 attributes giving information about the frequency of usage of some words, the frequency of capital letters and other insights to detect if the e-mail is a spam or not.
\item Waveform describes 3 types of waves with an added noise. Each class is generated from a combination of 2 of 3 "base" waves and each instance is generated of added noise (mean 0, variance 1) in each attribute.
\item  Wdbc {data is} Breast Cancer Wisconsin (Diagnostic) Data Set. Features are computed from a digitized image of a fine needle aspirate (FNA) of a breast mass. They describe characteristics of the cell nuclei present in the image. The target feature records the prognosis (benign (1) or malignant (2)). 
\item Wine {data is} the results of a chemical analysis of wines grown in the same region in Italy but derived from three different cultivars. The analysis determined the quantities of 13 constituents found in each of the three types of wines. 
\end{itemize}

 \textbf{\subsubsection{Data-set splitting}}
In order to test experimentally the proposed algorithm, we first proceeded with a data pre-processing in order to create the local subsets.

\begin{figure}[!h]
\centering
\includegraphics[width=0.70\textwidth]{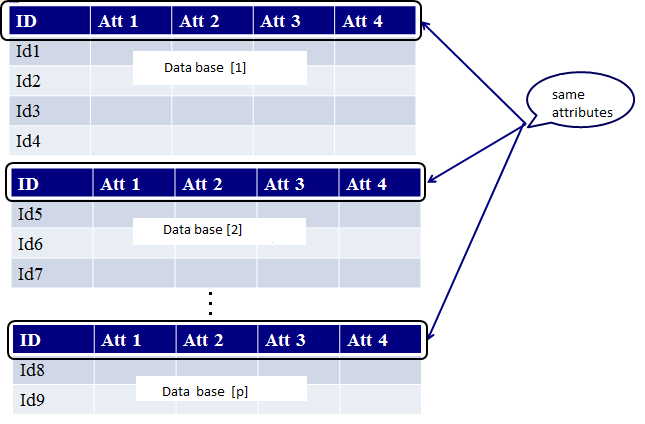}
\caption{Splitting  of the Vertical collaboration }
\label{ver}
\end{figure}

For vertical collaboration, we aim create samples from the original data, which means different instances represented with same characteristics. We split the data horizontally into 10 random subsets $X^{v}$, each subset is represented by the distribution $\mu^{v}$ we train the algorithm \ref{lmv} to get the local centroids partitions $\nu^{v}$, and then applied the collaborative algorithm \ref{collot} between the subsets in order to increase their local quality.  
To do so, we split the data as showed in figure \ref{ver} where the data base is rated into $v$ samples that share the same features. 

\begin{figure}[!h]
\centering
  \begin{center}
  \includegraphics[width=0.70\textwidth]{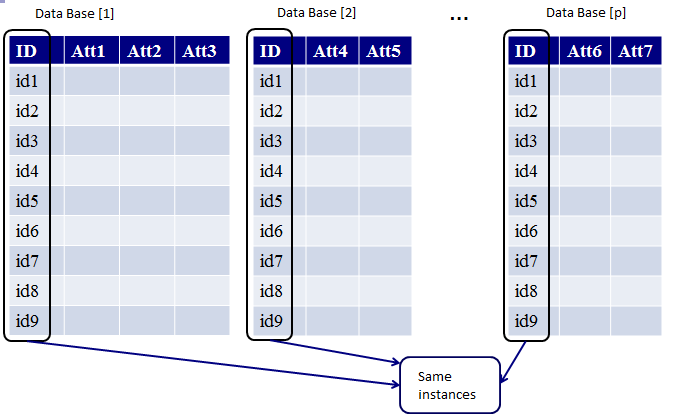}
  \end{center}
  \caption{Splitting  of the Horizontal collaboration }
\label{hor}
\end{figure}

For horizontal collaboration, the main idea is to split each chosen data set to 10 subsets, (see figure \ref{hor})
that share the same instances but represented with different features in each subset, selected randomly with replacement. Considering the notation above, each subset $X^{v}$ will be represented by the distribution $\mu^{v}$ that will be considered as the input of algorithm \ref{lmv} to get the distribution of the local centroids $\nu^{v}$. Algorithm \ref{collot} is then applied to influence the location of the local centroids by the centroids of the distant learners without having access to their local data.

 \textbf{\subsubsection{Quality measures}}
 The proposed approach was evaluated with two internal quality indexes: Davies-Bouldin (DB) and Silhouette indexes, as well as an external criterion: Adjusted Rand Index (ARI). 
$DB$-Davies Bouldin index \cite{davies1979cluster} is defined as follow:
\begin{equation}\label{Davi}
DB=\frac{1}{K}\sum_{k=1}^{K}\max_{k\neq {k'}} \frac{\Delta_n(c_{k})+\Delta_n(c_{k'})}{\Delta(c_{k},c_{k'})}
\end{equation}
Where $K$ is the number of clusters, $\Delta(c_k, c_{k'})$ is the similarity between clusters centers $c_k$ and $c_{k'}$ and $\Delta_n$ is the average similarity of all elements from the cluster $C_k$ to their cluster center $c_k$. 
This index evaluates the quality of unsupervised clustering basing on the compactness of clusters and separation measure between clusters. It's based on the ratio of the sum of within-clusters scatter to between-clusters separation. The lower the value of $DB$ index, the better the quality of the cluster.

The silhouette index \cite{rousseeuw1987silhouettes}, is based on the measurement of the difference between the average of the distance between the instance $x_{i}$ and the instances belonging to the same cluster $a_{i}$ and the average distance between the instance $x_{i}$ and the instances belonging to other clusters $b_{i}$, the closer the silhouette value is to 1 means that the instances are assigned to the right cluster.
\begin{equation}\label{Sil}
S=\frac{1}{K}\sum\frac{b(i)-a(i)}{\max(a(i),b(i))}
\end{equation}

Moreover, since the data-sets we proposed in the experiments provide available labels, we choose to add an external quality index the Adjusted Rand Index ($ARI$) \cite{steinley2004properties}.

The Adjusted Rand Index
\cite{wu2009adapting}  defined as follow \ref{ARI}:
\begin{equation}\label{ARI}
ARI=\frac{\sum_{ij}\binom{n_{ij}}{2}-\sum_{i}\binom{a_{i}}{2}\sum_{j}\binom{b_{j}}{2}/\binom{n}{2}}{\frac{1}{2}(\sum_{i}\binom{a_{i}}{2}+\sum_{j}\binom{b_{j}}{2})-\sum_{i}\binom{a_{i}}{2}\sum_{j}\binom{b_{j}}{2}/\binom{n}{2}}
\end{equation}
Where $n_{ij}=\mid C_{i}\cap Y_{j}\mid$ and $C_{i}$ is the $i$th cluster and $Y_{j}$ is $j$th  real class provides from the real label of the data-sets, and $a_{i}$ is the number of instances belonging to the same cluster with  the same class while $b_{j}$ is the number of instances belonging to different cluster with different class.  

The $ARI$ index measures the agreement between two partitions, one provided from the proposed algorithm and the second one provided from the labeled data-sets. The values of $ARI$ are between $0$ and $1$ and the quality is better when the value of $ARI$ is close to $1$.

 We therefore applied algorithm \ref{lmv} on local data, then the coefficient matrix $\alpha$ is computed based on a diversity index between the collaborators \cite{rastin2015collaborative}. This coefficient is used to control the importance of the terms of the collaboration. Algorithm \ref{collot} in trained 20 times in order to estimate the mean quality of the collaboration and a 95\% confidence interval for the 20 experiments.
The experimental results of horizontal collaboration were compared with SOM-collaborative \cite{ghassany2012collaborative}. Both approaches were trained on the same subsets and on the same local model, a $3\times 5$ map, with the parameters suggested by the authors of the algorithm \cite{ghassany2012collaborative}.
The last part of the experiments results consists to compare the proposed algorithm with the collaborative algorithms proposed in the state of the art, where the algorithms is trained on only two collaborators, we followed the same split as mentioned in \cite{ghassany2012collaborative} to compare the gain quality brought from the collaboration, based on $DB$ Davis-Bouldin index.

\textbf{\subsubsection{Computation tools}}
 
 A nice feature of the wasserstein distance is that their computation is vectored, which means the computation of a $n$ distances, whether from one histogram to many, or many to many, can be carried out simultaneously using elementary linear algebra operations. To do so we use the PyTorch version of Sinkhorn-means, on GPGPU’s. Moreover, the data collaborators were parallelized in order to compute local algorithm at the same time. For the experiment results, we used Alienware area-51m with GeForce RTX 2080/PCIe/SSE2 / NVIDIA Corporation graphic card.

\subsection{Results and discussion}

In this section we evaluate the approach on several data-sets for both vertical and horizontal clustering, based on different quality indexes, either internal or external one. We also compare the proposed algorithm with state-of-the-art approaches of collaborative clustering based on prototypes exchanges: Self-Organizing Maps collaboration ({Co-SOM}) and Generative-Topographic Maps collaboration ({Co-GTM}).

\subsubsection {Vertical Collaboration case}

To evaluate the proposed approach in a vertical collaboration case, we computed the algorithm \ref{collot} on several sub-sets that share the same features but have different size and complexity. 

\begin{table}[!h]
\caption{
Values of the different quality indexes before and after a vertical collaboration for each collaborator built from the Spambase data set.}
\label{vsp}
\centering
\setlength{\tabcolsep}{0.3cm}
\begin{tabular}[b]{l|cc|cc|cc}
\hline
\multirow{2}{*}{Models} 
&\multicolumn{2}{|c|}{DB} & \multicolumn{2}{|c|}{Silhouette}& \multicolumn{2}{|c}{ARI}\\
&\multicolumn{1}{|c|}{Before}& \multicolumn{1}{|c|}{After} & \multicolumn{1}{|c|}{Before}& \multicolumn{1}{|c|}{After} & \multicolumn{1}{|c|}{Before}& \multicolumn{1}{|c}{After}
\\\toprule
collab1  & 0.681 & 0.561 & 0.524 & 0.539 & 0.169 & 0.165 \\
collab2  & 0.769 & 0.540 & 0.532 & 0.625 & 0.118 & 0.119 \\
collab3  & 0.751 & 0.653 & 0.530 & 0.548 & 0.127 & 0.130 \\
collab4  & 0.714 & 0.576 & 0.538 & 0.574 & 0.168 & 0.169 \\
collab5  & 0.653 & 0.673 & 0.539 & 0.535 & 0.149 & 0.148 \\
collab6  & 0.714 & 0.569 & 0.552 & 0.556 & 0.156 & 0.160 \\
collab7  & 0.705 & 0.589 & 0.536 & 0.563 & 0.174 & 0.154 \\
collab8  & 0.720 & 0.576 & 0.544 & 0.590 & 0.163 & 0.165 \\
collab9  & 0.717 & 0.711 & 0.503 & 0.577 & 0.149 & 0.178 \\
collab10 & 0.665 & 0.605 & 0.495 & 0.561 & 0.159 & 0.192\\
\bottomrule
\end{tabular}
\end{table}

\begin{table}[!ht]
\centering
\caption{
Average values ($\pm CI_{95\%}$) of the different quality indexes before and after the vertical  collaboration for each data set over 20 executions.}  
\label{vic}
\renewcommand{\arraystretch}{1.2}
 \setlength{\tabcolsep}{0.1cm}
\begin{tabular}{l|c|cc|cc|cc}
\hline
\textbf{Indexes} & &\textbf{DB }& &\textbf{Silhouette} &   &\textbf{ARI}\\ \toprule
\textbf{Data-sets}&  & \textbf{before} & \textbf{after} &\textbf{before} & \textbf{after} & \textbf{before} & \textbf{after}\\ \hline
\multirow{2}{*}{\textbf{Glass}} & Average  & 0.984 & 0.689 & 0.369 & 0.471 & 0.223 & 0.244 \\
                 & $\pm CI_{95\%}$ &$\pm$0.09 & $\pm$0.17 & $ \pm $0.04 & $\pm $0.06 & $\pm $0.05 & $ \pm $0.07 \\
\multirow{2}{*}{\textbf{Spambase}}  & Average  &   0.711  & 0.603 & 0.529 & 0.567 & 0.153 & 0.158
\\
                 & $\pm CI_{95\%}$   & $ \pm $0.02 & $\pm$0.03 & $\pm $0.01 & $\pm $0.01 & $\pm $0.01 & $ \pm $0.01
 \\
            
\multirow{2}{*}{\textbf{\parbox{0.1\linewidth}{Waveform-noise}}} & Average  & 2.819 & 2.768 & 0.078 & 0.080 & 0.285 & 0.291 \\
                 & $\pm CI_{95\%}$   & $\pm$0.05 & $\pm$0.05 & $\pm$0.002 & $\pm$0.002 & $\pm$0.01 & $\pm$0.01 \\
\multirow{2}{*}{\textbf{WDBC}} & Average  &  0.675 & 0.629 & 0.448 & 0.513 & 0.290 & 0.374
\\
                 & $\pm CI_{95\%}$   & $ \pm $0.02 & $\pm$0.05 & $ \pm $0.02 & $ \pm $0.02 & $ \pm $0.03 & $ \pm $0.06
\\
                 
\multirow{2}{*}{\textbf{Wine}} & Average  &  0.525 & 0.496 & 0.568 & 0.574 & 0.306 & 0.308

\\
                 & $\pm CI_{95\%}$   & $ \pm $0.02 & $\pm$0.02 & $\pm $0.02 & $\pm $0.02 & $\pm $0.03 & $ \pm $0.02
\\
\bottomrule
\end{tabular}
\end{table}

As one can see, the proposed approach shows in general an acceptable capacity at improving the $DB$ index of the clustering before and after a vertical collaboration table \ref{vic}. This is not surprising, considering that the proposed algorithm evaluates the gain of quality based on this index. The $DB$ index is computed at each iteration in order to learn whether or not the collaborator can benefit from this collaboration. To make sure of the validity of the algorithm, we used the silhouette internal index.
As shown in Table \ref{vic}, the value of Silhouette increases after collaboration, which is a confirmation that the proposed approach increases the quality of each collaborator. However, the quality gain resulting from the collaboration is not always very high for some data-sets. This is due to the structure of the database and its horizontal splitting. If the data is very sparse (notably Spambase), we can observe that the collaboration increases more the quality than non-sparse data (for example Waveform data-set).

Table \ref{vic} shows the results achieved on this index and highlighted the performance of our algorithm and confirm that the quality of each collaborator increases after the collaboration. 

As one can see, the results are generally positive but the difference between the values either in internal indexes ($Silhouette$ and $DB$) or in external index $ARI$ before and after collaboration is not very impressive, this is explained by the horizontal splitting which gives small subsets that practically have the same structure, which means that the collaboration can be seen as a bidirectional exchange of information between subsets of the same given database. 

As we will see later on, this is not the case in horizontal collaboration, in which the impact of the collaboration is more important since the data are represented with different features for each collaborator. 
In addition, we chose one data set (due to page limitation) to detail the effect of the proposed algorithm on each collaborator. Table \ref{vsp} shows the values of different quality indexes of each collaborator built from Spambase data set, and confirm that the quality does increase the quality of most collaborators in the process.

\begin{figure}[!h]
 \centering
 \begin{subfigure}{.3\textwidth}
 \centering
 \includegraphics[width=1\linewidth]{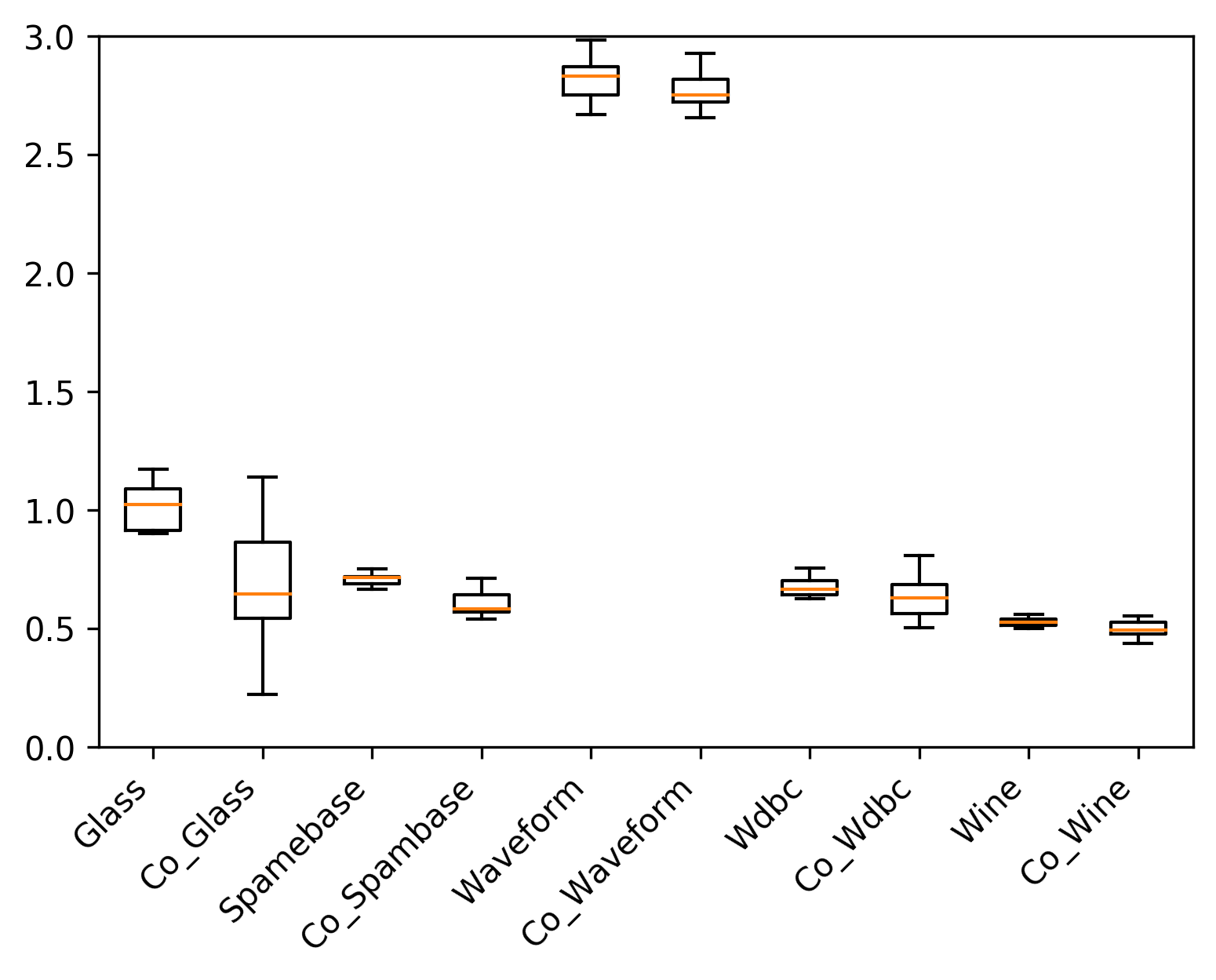} \caption{Davis Bouldin}
 \end{subfigure}
 \begin{subfigure}{.3\textwidth}
 \centering
 \includegraphics[width=1\linewidth]{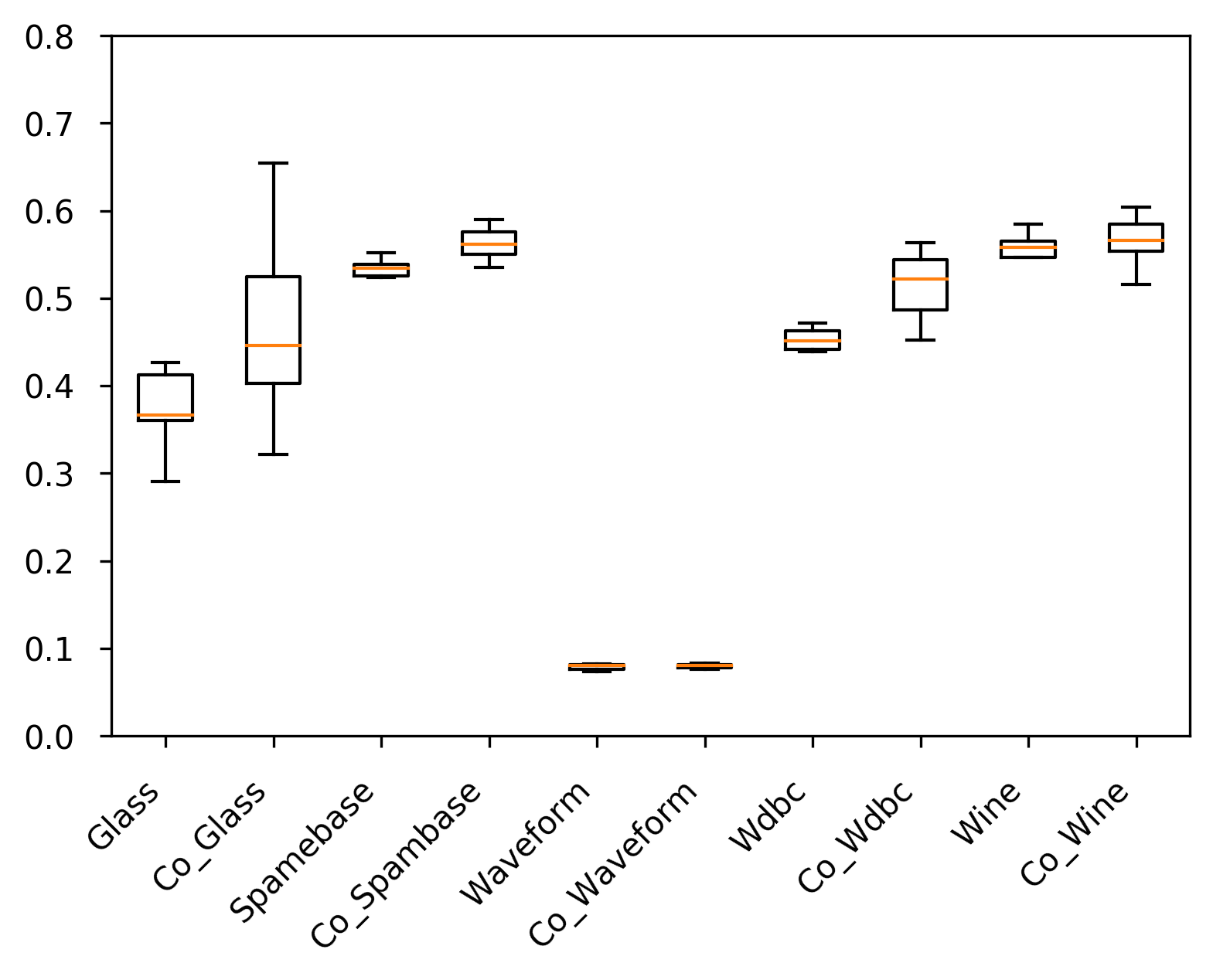} \caption{Silhouette }
 \end{subfigure}
 \begin{subfigure}{.3\textwidth}
 \centering
 \includegraphics[width=1\linewidth]{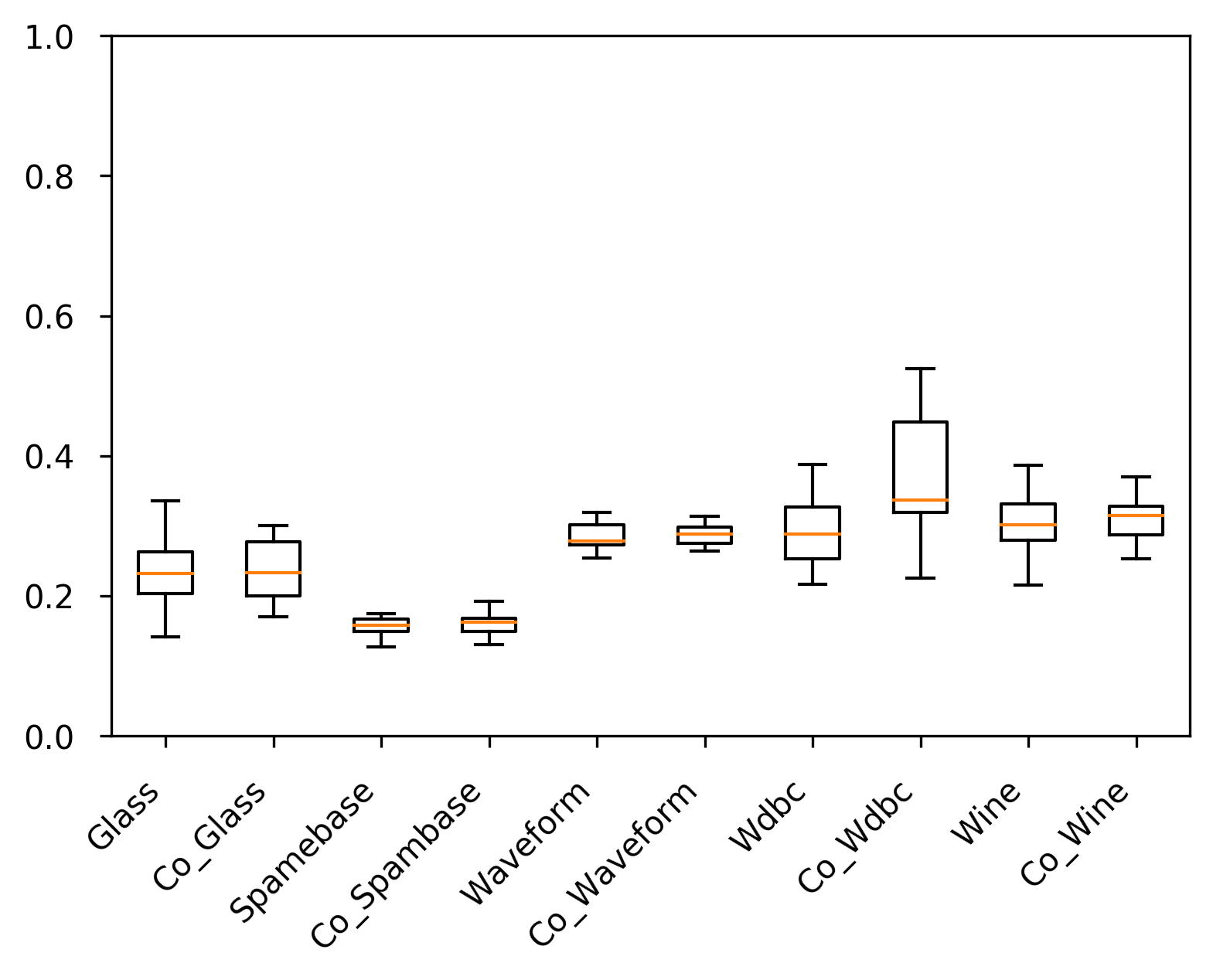}\caption{ARI}
 \end{subfigure} 

\caption{Sensitivity Box-Whiskers plots for the vertical collaboration case}
 \label{vboxes}
 \end{figure}
 
 Sensitivity Box-Whiskers plots (figure \ref{vboxes}) are drawn for the 20 experiments scores for each dataset before and after collaboration process. They enable us to study the distributional characteristics of scores as well as the level of the scores. To begin with, scores are sorted over the 20 tests. Then four equal sized groups are made from the ordered scores. That is, 25\% of all scores are placed in each group. The lines dividing the groups are called quartiles, and the groups are referred to as quartile groups. Usually, we label these groups 1 to 4 starting at the bottom. The median (middle quartile) marks the mid-point of the scores and is shown by the line that divides the box into two parts. Half the scores are greater than or equal to this value and half are less. The middle “box” represents the middle 50\% of scores for the group. The range of scores from lower to upper quartile is referred to as the inter-quartile range. The middle 50\% of scores fall within the inter-quartile range.

As can be seen from these graphs, the overall performance behavior shows a clear improvement as a result of the collaboration process. For example, for the DB Index, we can see a decrease in index values for all databases due to the contribution of collaboration. For the other two quality indices, we rather observe an increase in the values of the indices showing an improvement in the qualities of the solutions found.

\subsubsection{Horizontal collaboration case}

In this section we validate the effectiveness of the proposed approach on different date-sets for horizontal collaboration, where each collaborator represents the instances with different features (in a different representation space) see figure \ref{hor}.

We show how the exchange of information between the collaborators can improve the local results of each collaborator. Moreover, we show that the gain of quality is much important comparing to classical collaboration (SOM and GTM collaboration) 


Besides Davis-Bouldin index \ref{Davi}, which is trained in the algorithm, we validated the proposed approach with silhouette \ref{Sil}, the Adjusted Rand Index \ref{ARI}.

Table \ref{hic} shows that the collaboration step in the proposed approach increases the local quality of the models in regard to internal indexes $DB$ and $Silhouette$, in a horizontal collaboration framework, for different data-sets. Similarly, the $ARI$ index values show that the clusters computed by the models are closer to the expected output after the collaborations (table \ref{hic}).
One can notice that horizontal collaboration, between models that do not share the same representation space, is much more beneficial compared to vertical collaboration, where the models are computed in different spaces. This is due to the fact that in the vertical framework, the random splitting of the data-sets produce sub-sets of different instances represented in the same space (i.e., same features) with similar distributions due to the random process of the split. Therefore, each local model should be quite similar to the others and few exploitable information is exchanged in the collaborative step.
This could be confirmed by the comparison between the index values of Spambase data set in vertical collaboration (table \ref{vsp}) and the horizontal collaboration (table \ref{hsp}) where the difference between the score of the indexes is much more important for each collaborator comparing to vertical collaboration.



\begin{table}[!h]
\caption{
Values of the different quality indexes before and after the horizontal collaboration for each collaborator built from the Spambase data set.}
\label{hsp}
\centering
\setlength{\tabcolsep}{0.3cm}
\begin{tabular}[b]{l|cc|cc|cc}
\hline
\multirow{2}{*}{Models} 
&\multicolumn{2}{|c|}{DB} & \multicolumn{2}{|c|}{Silhouette}& \multicolumn{2}{|c}{ARI}\\
&\multicolumn{1}{|c|}{Before}& \multicolumn{1}{|c|}{After} & \multicolumn{1}{|c|}{Before}& \multicolumn{1}{|c|}{After} & \multicolumn{1}{|c|}{Before}& \multicolumn{1}{|c}{After}
\\\toprule
collab1  & 0,583 & 0.565  & 0.415 & 0.532 & 0.045 & 0.137\\
collab2  & 0.751 & 0.690  & 0.392 & 0.452  & 0.086 & 0.136\\
collab3  & 0.555 & 0.495  & 0.543 & 0.788 & 0.043 & 0.135\\
collab4  & 1.436 & 0.578   & 0.315 & 0.631 & 0.073 & 0.118\\
collab5  & 0.714 & 0.459   & 0.507 & 0.717 & 0.057 & 0.136 \\
collab6  & 1.067 & 0.706   & 0.287 & 0.578& 0.058 & 0.139 \\
collab7  & 1.183 & 1.099   & 0.304 & 0.312 & 0.157 & 0.144\\
collab8  & 0.722 & 0.511   & 0.505 & 0.470 & 0.101 & 0.143\\
collab9  & 0.707 & 0.503   & 0.435 & 0.555 & 0.036 & 0.136\\
collab10 & 1.370 & 0.418   & 0.202 & 0.755& 0.069 & 0.132\\
\bottomrule
\end{tabular}
\end{table}

\begin{table}[!h]
\centering
\caption{
 Average values ($\pm CI_{95\%}$) of the different quality indexes before and after the horizontal collaboration for each data set over 20 executions. } 
\label{hic}
\renewcommand{\arraystretch}{1.2}
 \setlength{\tabcolsep}{0.1cm}
\begin{tabular}{l|c|cc|cc|cc}
\hline
\textbf{Indexes} & &\textbf{DB }& &\textbf{Silhouette} &   &\textbf{ARI}\\ \toprule
\textbf{Data-sets}&  & \textbf{before} & \textbf{after} &\textbf{before} & \textbf{after} & \textbf{before} & \textbf{after}\\ \hline
\multirow{2}{*}{\textbf{Glass}} & Average  & 1.028 & 0.608 & 0.335 & 0.552 & 0.155 & 0.237  \\
                 & $\pm CI_{95\%}$ & $\pm$0.23 & $\pm$0.18 & $\pm$0.02 & $\pm$0.04 &$\pm$0.04 & $\pm$0.01 \\
\multirow{2}{*}{\textbf{Spambase}}  & Average  &  0.903  & 0.481 & 0.390 & 0.579 & 0.072 & 0.135
 \\
                 & $\pm CI_{95\%}$   & $\pm$0.20 & $\pm$0.12 & $\pm$0.06 & $\pm$0.09 & $\pm$0.02 & $\pm$0.004 \\
            
\multirow{2}{*}{\textbf{\parbox{0.1\linewidth}{Waveform-noise}}} & Average  & 2.578 & 2.310 & 0.078 & 0.108 & 0.179 & 0.218 \\
                 & $\pm CI_{95\%}$   & $\pm$0.15 & $\pm$0.20 & $\pm$0.008 & $\pm$0.01 & $\pm$0.02 & $\pm$0.02 \\
\multirow{2}{*}{\textbf{WDBC}} & Average  &  0.601& 0.550 & 0.483 & 0.566 & 0.219 & 0.439\\
                 & $\pm CI_{95\%}$   & $\pm$0.17 & $\pm$0.09 & $\pm$0.02 & $ \pm$0.05 & $\pm$0.05 & $\pm$0.13\\
                 
\multirow{2}{*}{\textbf{Wine}} & Average  &  0.688 & 0.643 & 0.470 & 0.490 & 0.206 & 0.212
\\
                 & $\pm CI_{95\%}$   & $ \pm$0.10 & $ \pm$0.09 & $ \pm$0.06 & $\pm$0.05 & $\pm$0.05 & $ \pm$0.05\\
\bottomrule
\end{tabular}
\end{table}

\begin{figure}[!h]
 \centering
 \begin{subfigure}{.3\textwidth}
 \centering
 \includegraphics[width=1\linewidth]{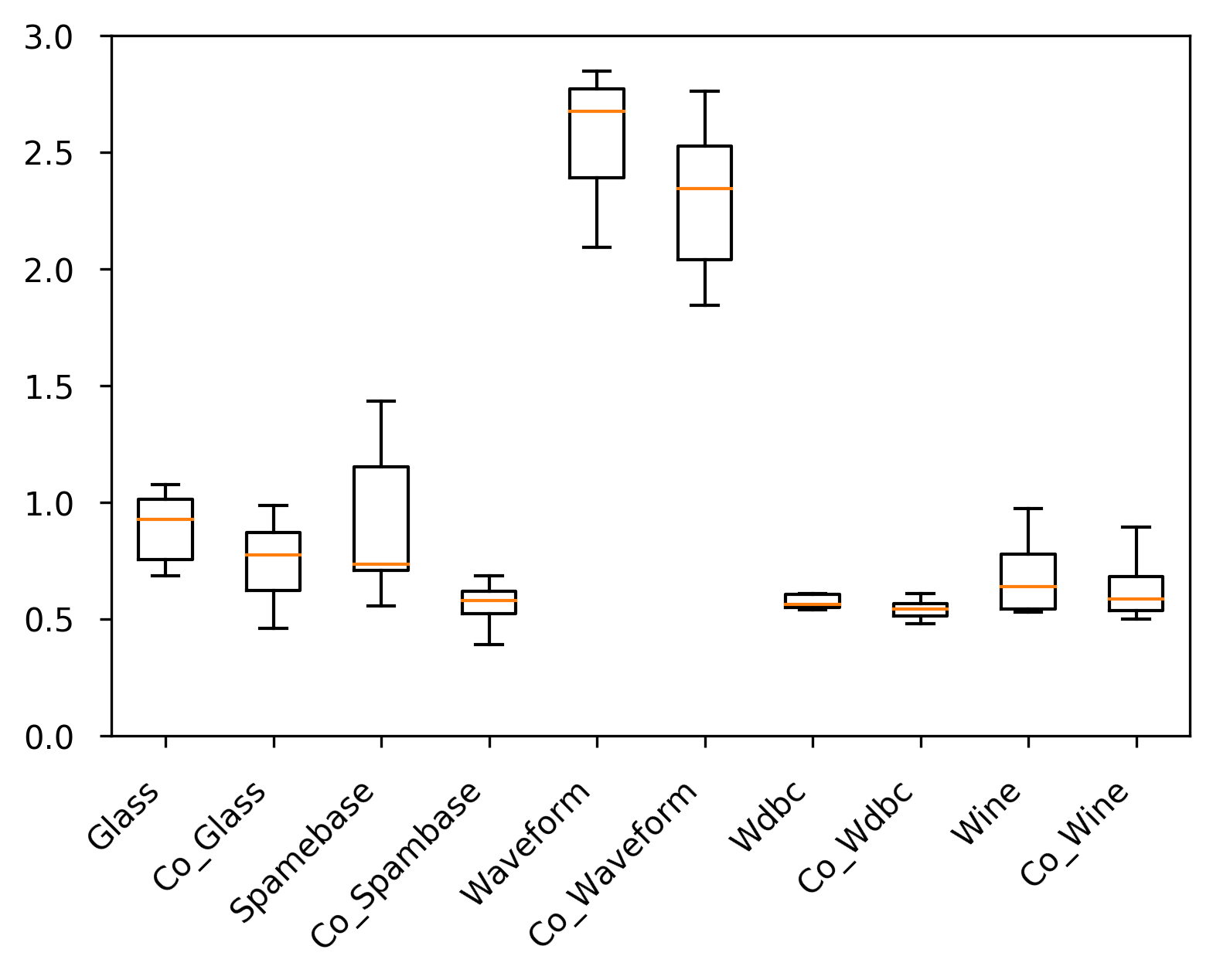} \caption{Davis Bouldin}
 \end{subfigure}
 \begin{subfigure}{.3\textwidth}
 \centering
 \includegraphics[width=1\linewidth]{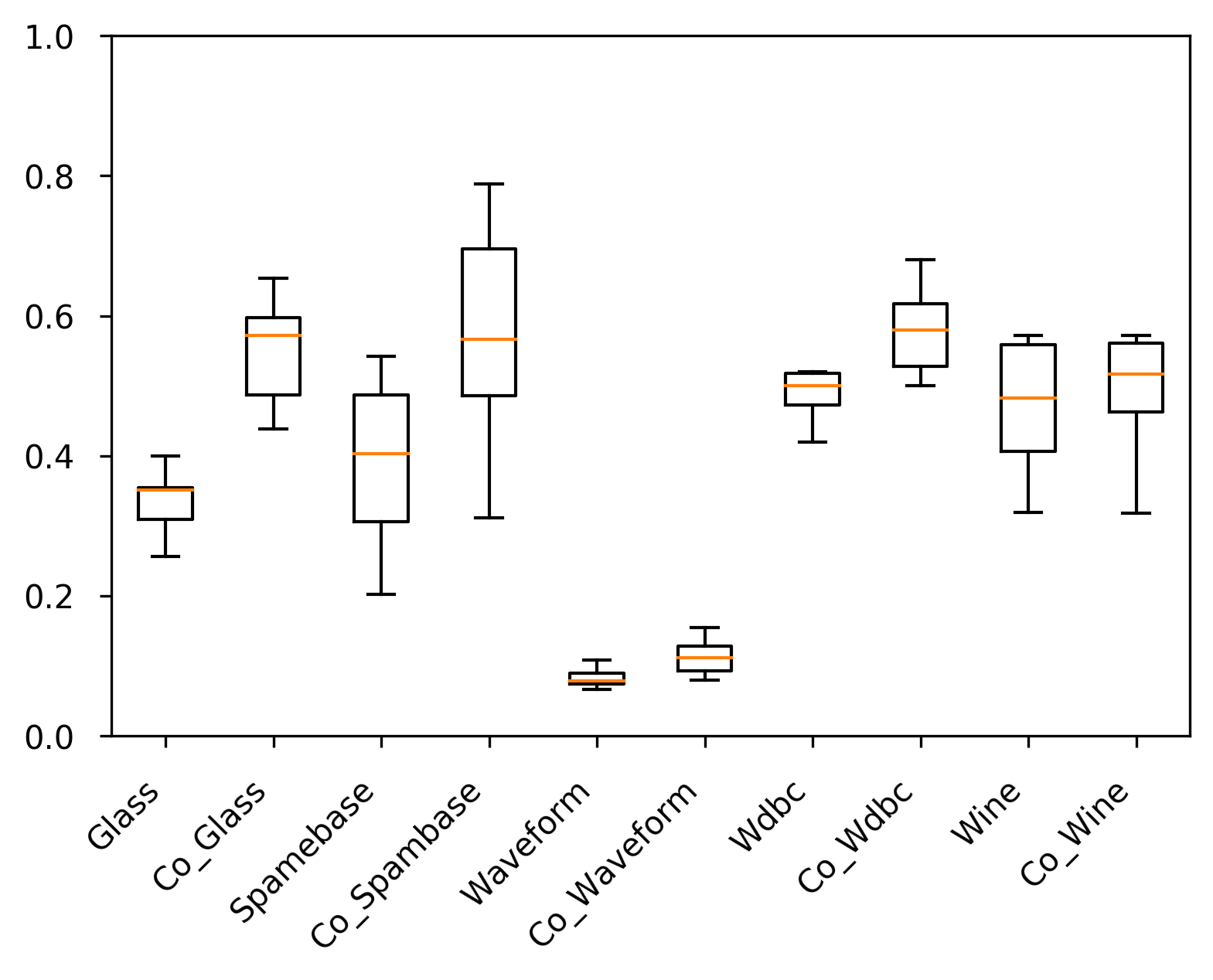} \caption{Silhouette }
 \end{subfigure}
 \begin{subfigure}{.3\textwidth}
 \centering
 \includegraphics[width=1\linewidth]{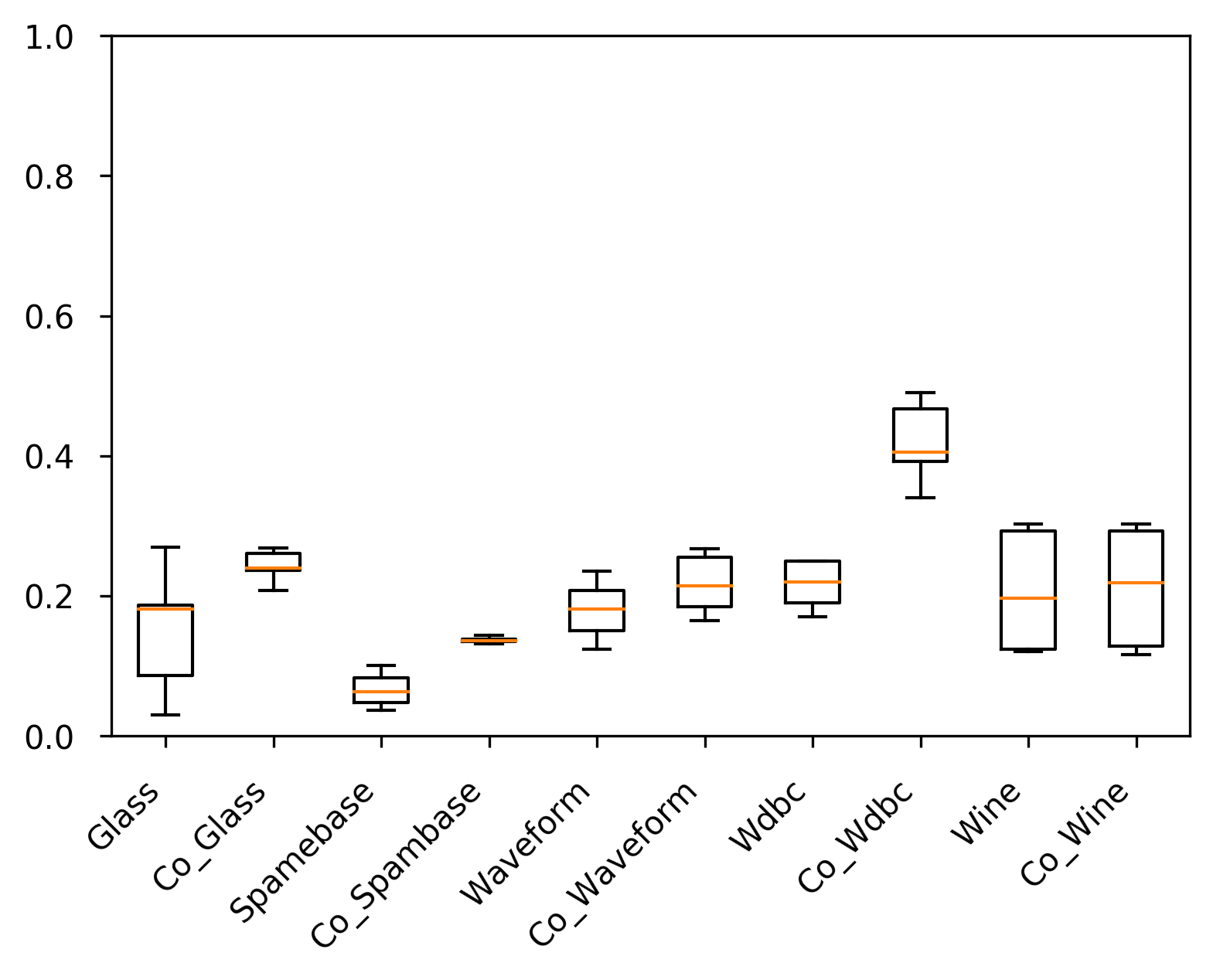}\caption{ARI}
 \end{subfigure} 

\caption{Sensitivity Box-Whiskers plots for the horizontal collaboration case}
 \label{hboxes}
 \end{figure}

 Sensitivity Box-Whiskers plots (figure \ref{hboxes}) represents a synthesis of the scores into five crucial pieces of information identifiable at a glance: position measurement, dispersion, asymmetry and length of Whiskers. The position measurement is characterized by the dividing line on the median (as well as the middle of the box). Dispersion is defined by the length of the Box-Whiskers (as well as the distance between the ends of the Whiskers and the gap). Asymmetry is defined as the deviation of the median line from the center of the Box-Whiskers from the length of the box (as well as by the length of the upper Whiskers from the length of the lower Whiskers, and by the number of scores on each side). The length of the Whiskers is the distance between the ends of the Whiskers in relation to the length of the Box-Whiskers (and the number of scores specifically marked).
These graphs show the same overall performance behavior observed in the case of vertical collaboration. They show a clear improvement as a result of the collaboration process. This improvement is observed for all quality indices used.

\subsubsection{Comparison with other collaborative approaches} 
\label{subcompa}

In this section, the proposed collaborative algorithm \ref{collot} is based on Sinkhorn-Means (Sin-Mean) local algorithms as described in algorithm \ref{lmv} (this framework is thereafter called {Co-Sin-OT}) and we illustrate the adaptability of the proposed collaborative approach by alternatively using Self-Organizing Maps (SOM) as local algorithms ({Co-SOM-OT}). Both are compared to popular state-of-the-art collaborative algorithms based on Self-Organized-Maps ({Co-SOM}) \cite{grozavu2010topological} and Generative-Topographic-Maps (Co-GTM) \cite{ghassany2012collaborative}. We focus here on the horizontal collaboration case as in \cite{grozavu2010topological} and  \cite{ghassany2012collaborative}. Indeed, horizontal collaboration is usually more useful and applicable to real problems comparing to vertical collaboration, it is also more difficult. In the first part of the experiments, we test the quality of the collaboration for 10 collaborators. As the Co-GTM algorithm is designed for only two collaborators, it is not included in the comparisons. In the second part, only two collaborators are trained and the Co-GTM algorithm is included in the protocol.  

\begin{table}[!h]
\caption{ Comparison of SOM-based and Sinkhorn-based collaborative approaches, using the $Silhouette$ index for the Glass data set. The average values ($\pm CI_{95\%}$) is computed over 20 executions.}

\label{COM1}
\centering
\resizebox{\textwidth}{!}{
\begin{tabular}{l|ccc|ccccc}
\hline
&\multicolumn{3}{|c|}{Sinkhorn-based}&\multicolumn{5}{|c|}{SOM-based} \\
&\multicolumn{1}{|c|}{Sin-Means}&\multicolumn{1}{|c|}{Co-Sin-OT}& \multicolumn{1}{|c|}{{Gain}}&\multicolumn{1}{|c|}{SOM}& \multicolumn{1}{|c|}{{ Co-SOM}}&\multicolumn{1}{|c}{{Gain}}& \multicolumn{1}{|c|}{{Co-SOM-OT}}&\multicolumn{1}{|c|}{{Gain}}  \\\toprule
collab1  & 0.353 & 0.582 & 0.229 & 0.088  & 0.240 & 0.152 & 0.240 & 0.152  \\
collab2  & 0.294 & 0.461 & 0.167 & -0.009 & -0.009  & 0.000 & 0.131 & 0.140  \\
collab3  & 0.351 & 0.653 & 0.303 & -0.036 & -0.036  & 0.000 & 0.156 & 0.192  \\
collab4  & 0.400 & 0.569 & 0.169 & 0.395  & 0.395   & 0.000 & 0.340 & -0.055 \\
collab5  & 0.256 & 0.439 & 0.182 & -0.008 & 0.320   & 0.329 & 0.320 & 0.329  \\
collab6  & 0.339 & 0.602 & 0.263 & 0.070  & 0.070   & 0.000 & 0.102 & 0.032  \\
collab7  & 0.299 & 0.565 & 0.266 & 0.222  & 0.257   & 0.034 & 0.253 & 0.031  \\
collab8  & 0.358 & 0.444 & 0.086 & 0.410  & 0.410   & 0.000 & 0.439 & 0.029  \\
collab9  & 0.351 & 0.635 & 0.284 & 0.073  & 0.183   & 0.110 & 0.223 & 0.150  \\
collab10 & 0.355 & 0.574 & 0.220 & -0.053 & -0.051  & 0.001 & 0.003 & 0.056\\
\hline
Average & 0.335 & 0.552 & 0.216 & 0.115 & 0.177 & 0.063 & 0.221 & 0.106 \\ 
$\pm CI_{95\%}$ & $\pm$0.02 & $\pm$0.04 & $\pm$0.12 &  $\pm$0.10 & $\pm$0.10 &  $\pm$0.06 & $\pm$0.07 & $\pm$0.06 \\
\bottomrule
\end{tabular}}
\end{table}

 The first set of experiments are thus restricted to {Co-SOM}, {Co-SOM-OT} and {Co-Sin-OT} in order to be able to work with several collaborators. All collaborative approaches are applied on the same subsets. In SOM-based approaches, each local collaborator starts with the same $5\times3$ SOM. The approaches are compared using the Silhouette index. As shown in Tables \ref{COM1} to \ref{COM5}, the results obtained with the proposed approach are globally better for this index. One can note that, for some collaborators, the quality of the collaboration leads to very similar results in both cases, despite very different quality before collaboration. The OT-based approach ({Co-SOM-OT}) provides a much more stable quality improvement over the set of collaborators. In addition, the use of Sinkhorn-Means as the local algorithm ({Co-Sin-OT}) provide the best results comparing a SOM-Based local clustering ({Co-SOM} and {Co-SOM-OT}). This can be explained by the fact that the mechanism of the SOM-based collaborative algorithms is constrained by the neighborhood’s functions. Moreover, it was built for a collaboration between two collaborators, then extended to allows multiple collaborations, unlike the proposed approach where each learner exchange information with all of the others at each step of the collaboration.

\begin{table}[!h]
\caption{  Comparison of SOM-based and Sinkhorn-based collaborative approaches, using the $Silhouette$ index for the Spambase data set. The average values ($\pm CI_{95\%}$) is computed over 20 executions.}

\label{COM2}
\centering
\resizebox{\textwidth}{!}{
\begin{tabular}{l|ccc|ccccc}
\hline
&\multicolumn{3}{|c|}{Sinkhorn-based}&\multicolumn{5}{|c|}{SOM-based} \\
&\multicolumn{1}{|c|}{Sin-Means}&\multicolumn{1}{|c|}{Co-Sin-OT}& \multicolumn{1}{|c|}{{Gain}}&\multicolumn{1}{|c|}{SOM}& \multicolumn{1}{|c|}{{ Co-SOM}}&\multicolumn{1}{|c}{{Gain}}& \multicolumn{1}{|c|}{{Co-SOM-OT}}&\multicolumn{1}{|c|}{{Gain}}  \\\toprule
collab1  & 0,415 & 0,532 & 0,118  & 0,224  & 0,483 & 0,260 & 0,346  & 0,122 \\
collab2  & 0,392 & 0,452 & 0,060  & 0,038  & 0,080   & 0,042 & 0,124  & 0,086 \\
collab3  & 0,543 & 0,788 & 0,246  & -0,137 & -0,137  & 0,000 & 0,005  & 0,142 \\
collab4  & 0,315 & 0,631 & 0,316  & -0,308 & -0,091  & 0,216 & -0,103 & 0,205 \\
collab5  & 0,507 & 0,717 & 0,211  & -0,101 & -0,028  & 0,073 & 0,052  & 0,153 \\
collab6  & 0,287 & 0,578 & 0,291  & -0,039 & 0,036   & 0,075 & 0,153  & 0,192 \\
collab7  & 0,304 & 0,312 & 0,008  & -0,035 & -0,035  & 0,000 & 0,087  & 0,122 \\
collab8  & 0,505 & 0,470 & -0,035 & 0,314  & 0,524   & 0,210 & 0,511  & 0,197 \\
collab9  & 0,435 & 0,555 & 0,119  & -0,260 & -0,069  & 0,191 & 0,023  & 0,283 \\
collab10 & 0,202 & 0,755 & 0,553  & 0,041  & 0,041   & 0,000 & 0,114  & 0,073\\
\hline
Average &0.390 & 0.579 & 0,188 &-0.026 &0.035 & 0.106 &0,131 & 0,158\\
$\pm CI_{95\%}$& $\pm$0.06&$\pm$0.09 &$ \pm$0.10&  $\pm$0.12&$ \pm$0.12 &  $\pm$0.06 &$ \pm$0.11 & $\pm$0.03\\
\bottomrule
\end{tabular}}
\end{table}

\begin{table}[!h]
\caption{ Comparison of SOM-based and Sinkhorn-based collaborative approaches, using the silhouette index for the Waveform-noise data set. The average values ($\pm CI_{95\%}$) is computed over 20 executions.}
\label{COM3}
\centering
\resizebox{\textwidth}{!}{
\begin{tabular}{l|ccc|ccccc}
\hline
&\multicolumn{3}{|c|}{Sinkhorn-based}&\multicolumn{5}{|c|}{SOM-based} \\
&\multicolumn{1}{|c|}{Sin-Means}&\multicolumn{1}{|c|}{Co-Sin-OT}& \multicolumn{1}{|c|}{{Gain}}&\multicolumn{1}{|c|}{SOM}& \multicolumn{1}{|c|}{{ Co-SOM}}&\multicolumn{1}{|c}{{Gain}}& \multicolumn{1}{|c|}{{Co-SOM-OT}}&\multicolumn{1}{|c|}{{Gain}}  \\\toprule
collab1  & 0,108 & 0,155 & 0,047 & 0,025 & 0,030 & 0,006 & 0,064 & 0,039\\
collab2  & 0,074 & 0,097 & 0,023 & 0,036 & 0,036 & 0,000 & 0,062 & 0,026 \\
collab3  & 0,075 & 0,091 & 0,016 & 0,070 & 0,070 & 0,000 & 0,069 & -0,001\\
collab4  & 0,067 & 0,088 & 0,021 & 0,043 & 0,047 & 0,003 & 0,064 &0,021\\
collab5  & 0,081 & 0,127 & 0,046 & 0,054 & 0,058 & 0,004 & 0,069 &  0,015\\
collab6  & 0,067 & 0,079 & 0,012 & 0,063 & 0,063 & 0,000 & 0,067 & 0,004 \\
collab7  & 0,093 & 0,128 & 0,036 & 0,026 & 0,026 & 0,000 & 0,063 & 0,037 \\
collab8  & 0,095 & 0,140 & 0,045 & 0,040 & 0,044 & 0,004 & 0,067 & 0,027 \\
collab9  & 0,081 & 0,120 & 0,039 & 0,031 & 0,038 & 0,007 & 0,065 & 0,034\\
collab10 & 0,075 & 0,104 & 0,029 & 0,025 & 0,032& 0,007 & 0,063 & 0,038\\
\hline
Average &0.078 & 0.108 &0,0313 &0.043 &0.045 &	0.003 &0,065&0,024 \\ 
$\pm CI_{95\%}$ & $\pm$0.008 & $ \pm$0.01 & $\pm$0.008 & $\pm$0.01& $\pm$0.009 &$\pm$0.01 &$ \pm$0.05&$\pm$0.01\\
\bottomrule
\end{tabular}}
\end{table}

In the second set of experiments, we compare the proposed approach to classical collaborative algorithms based on Self-Organized-Maps ({Co-SOM}) \cite{grozavu2010topological} and Generative-Topographic-Maps ({Co-GTM}) \cite{ghassany2012collaborative}. The three approaches are compared using $DB$ index, as in \cite{ghassany2012collaborative, grozavu2010topological}. As shown in Table \ref{com}, the results obtained with the proposed approach ({Co-Sin-OT}), in comparison to the state-of-the-art, is generally better than the classical approaches. The lowest qualities are expressed by the older approach, SOM-based collaborative clustering ({Co-SOM}), followed by the GTM-based approach ({Co-GTM}). Unlike {Co-SOM} and {Co-GTM}, the proposed approach aims to find a local optimum for each collaborator. More precisely, at the end of the local training, each collaborator exchange information based on a stopping criterion that ends the collaboration with each collaborator as soon as the quality of the collaboration starts decreasing, which is not the case in the other approaches. Furthermore, Table \ref{com} compares the quality gain brought by the collaboration from each approach. The proposed approach increases the quality of each collaborator on all of the data-sets, which implies a positive gain quality. On the contrary, in SOM-based collaboration the gain can be negative for some data-sets. Finally, in order to evaluate the general performance of the approaches, we define the following score measurement:

\begin{equation}\label{score}
Score(M_i)= \sum _{j} \frac{G(M_i,D_j)} {\max_{i}G(M_i,D_j)} 
\end{equation}

Where $G$ indicates the gain quality of each approach $M_{i}$ of each data-sets $D_{j}$. This score gives an overall vision of the best approach on all the data-sets. As shown in table \ref{com}, the best score belongs to the proposed collaboration based on Optimal Transport theory, followed by the GTM collaborative approach ({Co-GTM}) and the SOM-based collaboration ({Co-SOM}). These results highlight the the performance of the proposed algorithm, due to the strong theoretical back-ground of Optimal Transport theory.

\begin{table}[!h]
\caption{Comparison of SOM-based and Sinkhorn-based collaborative approaches, using the silhouette index for the WDBC data set. The average values ($\pm CI_{95\%}$) is computed over 20 executions.}

\label{COM4}
\centering
\resizebox{\textwidth}{!}{
\begin{tabular}{l|ccc|ccccc}
\hline
&\multicolumn{3}{|c|}{Sinkhorn-based}&\multicolumn{5}{|c|}{SOM-based} \\
&\multicolumn{1}{|c|}{Sin-Means}&\multicolumn{1}{|c|}{Co-Sin-OT}& \multicolumn{1}{|c|}{{Gain}}&\multicolumn{1}{|c|}{SOM}& \multicolumn{1}{|c|}{{ Co-SOM}}&\multicolumn{1}{|c}{{Gain}}& \multicolumn{1}{|c|}{{Co-SOM-OT}}&\multicolumn{1}{|c|}{{Gain}}  \\\toprule
collab1  & 0,470 & 0,632 & 0,162 & 0,233 & 0,233 & 0,000 & 0,244 & 0,011 \\
collab2  & 0,514 & 0,620 & 0,106 & 0,185 & 0,208 & 0,024 & 0,211 & 0,026 \\
collab3  & 0,485 & 0,614 & 0,129 & 0,246 & 0,303 & 0,057 & 0,401 & 0,155 \\
collab4  & 0,521 & 0,608 & 0,087 & 0,015 & 0,074 & 0,059 & 0,126 & 0,111 \\
collab5  & 0,422 & 0,499 & 0,077 & 0,102 & 0,182 & 0,080 & 0,341 & 0,239 \\
collab6  & 0,490 & 0,549 & 0,059 & 0,240 & 0,278 & 0,038 & 0,334 & 0,094 \\
collab7  & 0,516 & 0,685 & 0,168 & 0,298 & 0,337 & 0,039 & 0,445 & 0,147 \\
collab8  & 0,388 & 0,392 & 0,004 & 0,125 & 0,125 & 0,000 & 0,323 & 0,198 \\
collab9  & 0,518 & 0,516 & -0,003 & 0,202 & 0,213 & 0,011 & 0,367 & 0,165 \\
collab10 & 0,513 & 0,548 & 0,035 & 0,110 & 0,123  & 0,013 & 0,236 & 0,126   \\
\hline
Average &0.483 & 0.566 &0,082 &0.175 &0.204 & 0.032 &0,303  & 0,127 \\
$\pm CI_{95\%}$&$\pm$0.02& $\pm$0.05 & $\pm$0.03 &$ \pm$0.05  &$\pm$0.05& $\pm$0.01 & $\pm$0.06& $\pm$0.04 \\
\bottomrule
\end{tabular}}
\end{table}

\begin{table}[!h]
\caption{Comparison of SOM-based and Sinkhorn-based collaborative approaches, using the silhouette index for the Wine data set. The average values ($\pm CI_{95\%}$) is computed over 20 executions.}

\label{COM5}
\centering
\resizebox{\textwidth}{!}{
\begin{tabular}{l|ccc|ccccc}
\hline
&\multicolumn{3}{|c|}{Sinkhorn-based}&\multicolumn{5}{|c|}{SOM-based} \\
&\multicolumn{1}{|c|}{Sin-Means}&\multicolumn{1}{|c|}{Co-Sin-OT}& \multicolumn{1}{|c|}{{Gain}}&\multicolumn{1}{|c|}{SOM}& \multicolumn{1}{|c|}{{ Co-SOM}}&\multicolumn{1}{|c}{{Gain}}& \multicolumn{1}{|c|}{{Co-SOM-OT}}&\multicolumn{1}{|c|}{{Gain}}  \\\toprule
collab1  & 0.560 & 0.562 & 0.002  & 0.3921 & 0.4033 & 0.011 & 0.446 & 0.054 \\
collab2  & 0.559 & 0.560 & 0.001  & 0.4288 & 0.4288 & 0.000 & 0.431 & 0.002 \\
collab3  & 0.572 & 0.573 & 0.001  & 0.4945 & 0.4945 & 0.000 & 0.542 & 0.048 \\
collab4  & 0.320 & 0.318 & -0.001 & 0.1223 & 0.1255 & 0.003 & 0.224 & 0.102 \\
collab5  & 0.446 & 0.499 & 0.054  & 0.1444 & 0.2116 & 0.067 & 0.221 & 0.077 \\
collab6  & 0.394 & 0.450 & 0.056  & 0.1558 & 0.1768 & 0.021 & 0.201 & 0.045 \\
collab7  & 0.329 & 0.351 & 0.022  & 0.1107 & 0.1107 & 0.000 & 0.257 & 0.146 \\
collab8  & 0.520 & 0.534 & 0.014  & 0.1243 & 0.2210 & 0.097 & 0.348 & 0.224 \\
collab9  & 0.444 & 0.498 & 0.054  & 0.0715 & 0.2286 & 0.157 & 0.356 & 0.284 \\
collab10 & 0.559 & 0.560 & 0.002  & 0.3923 & 0.3923 & 0.000 & 0.395 & 0.003\\
\hline
Average &0.470&	0.490&	0.020&	0.244&	0.279&	0.036&	0.342&	0.098 \\
$\pm CI_{95\%}$&$\pm$0.06	&$\pm$0.05 &$\pm$0.01	&$\pm$0.10	&$\pm$0.09	&$\pm$0.03	&$\pm$0.07	&$\pm$0.06\\

\bottomrule
\end{tabular}}
\end{table}

\begin{table}[!h]
\centering
\caption{Comparison of $DB$ index Between SOM, GTM and OT based approaches on different data-sets for two collaborators}
\label{com}
\renewcommand{\arraystretch}{1.2}
 \setlength{\tabcolsep}{0.01cm}
\begin{tabular}{lc|ccc|ccc|ccc}
\hline
\textbf{Approaches} & &\textbf{Co-SOM }&& &\textbf{Co-GTM} &  & &\textbf{Co-Sin-OT}\\ \toprule
\textbf{Data-sets}&  & \textbf{before} & \textbf{after} &\textbf{ gain}&\textbf{before} & \textbf{after} &\textbf{ gain} & \textbf{before} & \textbf{after} &\textbf{ gain}\\ \hline
\multirow{2}{*}{\textbf{Glass}}      & {Collab1}     & 1.010                  & 0.985  &  \multirow{2}{*}{0.002
}  &0.740 & 0.970     &  \multirow{2}{*}{0.000}                                     &1.109                   & 0.774 &      \multirow{2}{*}{0.253}                 \\
                   & {Collab2}    & 0.902                 & 0.924& &1.280  &1.050& &0.908   & 0.731&\\\hline
\multirow{2}{*}{\textbf{Spambase}}  & {Collab1}     & 2.924                  & 2.436& \multirow{2}{*}{-0.077} &1.120&1.060     &  \multirow{2}{*}{1.014}                                          &1.467                   & 1.055 &    \multirow{2}{*}{  0.187 }                 \\
& {Collab2}     & 0.960                 & 1.748 & &0.870&0.900& &0.775                  & 0.766 &\\\hline
            
\multirow{2}{*}{\textbf{\parbox{0.1\linewidth}{Waveform-noise}}} & { Collab1}    & 6.488& 6.488  &\multirow{2}{*}{0.027} & 1.140 &1.310 &  \multirow{2}{*}{0.378}                                            &3.592                   & 2.579    & \multirow{2}{*}{0.256}                   \\
& {Collab2}     & 7.269                  & 6.898     & &3.750&1.310& &3.732                   & 2.868 & \\\hline
\multirow{2}{*}{\textbf{Wdbc}} & {Collab1}     & 0.640   & 0.641 &     \multirow{2}{*} {0.001} &0.970 & 0.920 & \multirow{2}{*}{0.010}    &0.755                   &  0.612  &          \multirow{2}{*} {0.219}        \\
& {Collab2}     & 0.651               & 0.649&  & 0.870&0.900& &0.928                   & 0.701 &\\
\hline
\hline
\textbf{Score}  &   &&&\textbf{-1.740}   &&&\textbf{1.377}   &&&\textbf{3.581}\\\bottomrule
\end{tabular}
\end{table}

\begin{figure} [!h]
    
    \centering
    \includegraphics[scale = 0.35]{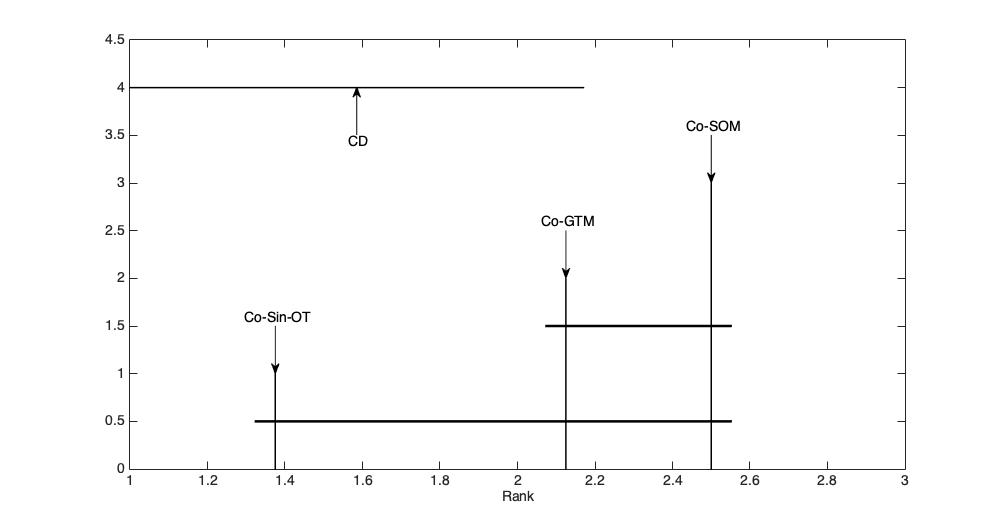}
   \caption{ Friedman and Nemenyi test for comparing multiple approaches over multiple data sets: Approaches are ordered from left (the best) to right (the worst)}  
   \label{fig:FriedmanTest}
\end{figure}

In order to assess the performance of our approaches, we use the Friedman test and Nemenyi test recommended in \cite{Demsar06}. The Friedman test is conducted to test the null-hypothesis that all approaches are equivalent in respect of accuracy. If the null hypothesis is rejected, then the Nemenyi test will be performed. In addition, if the average ranks of two approaches differ by at least the critical difference (CD), then it can be concluded that their performances are significantly different. In the Friedman test, we set the significant level $ \alpha=0.05$. The figure \ref{fig:FriedmanTest} shows a critical diagram representing a projection of average ranks approaches on enumerated axis. The approaches are ordered from left (the best) to right (the worst) and a thick line which connects the approaches were the average ranks not significantly different (for the level of 5\% significance). As shown in figure \ref{fig:FriedmanTest}, Co-Sin-OT achieves significant improvement over the other proposed techniques (Co-GTM and Co-SOM) since during collaboration phase it is stable and the process stops the collaboration for some learners when their local quality stars to decrease, which prevents common issue of collaborative approaches.

Compared to the most cited approaches of the state of the art, the positive impact of using the collaborative learning based on this theory is:
\begin{itemize}

 \item  The proposed algorithm is based on a strong and well defended theory that becomes increasingly popular in the field of machine learning. 

 \item  Its strength is highlighted by experimental validation on both for artificial and real data-sets. 

 \item The stopping criteria that we proposed based on the measure of the gain quality brought  after  each collaboration, because it  guarantee the convergence once the gain quality  tends towards zero.

 \item The choice of the distant collaborator which is very important and allows to give an optimal order of the collaboration. In the proposed algorithm we solved this paradigm based on the Optimal Transport Matrix plan, that aims to compare all distribution of the centroids in each site. In this way each collaborator will be enable to choose the best one. 

 \item The proposed algorithm stops the negative collaboration, based on the measure of the gain quality between each collaboration and updates the centroids if the gain quality is positive. Otherwise, it moves to the other distant collaborator.

\end{itemize}

Finally, the proposed approach ensures the adaptability of working with different local models, this is lead us to introduce some managerial applications of our work, like the management system learning where the collaborative learning could offer in interaction between learners to make them work cooperatively rather than competitively and helps to create sub-networks of collaboration where the diversity is decreased, and manage the conflict learning by using a one-to-one collaboration. Besides the exchange information using the proposed algorithm preserve the privacy of each collaborator, and ensures the control of the shared information with each collaborator, and filters the received information to avoid affecting the real structure of local data. Thus, all the collaborators can explore the distributed data that could containing some mutual information while keeping the control on received and transmitted information.

However, the proposed algorithm still suffers from some limitations, in particular considering the same dimension in every site, and also the curse of height dimensionality that Optimal Transport still suffers from, which leads us to increase the penalty coefficient of the regularization in order to avoid the over-fitting.

\section{Conclusion}
\label{sec:Conclusion}

In this paper, we proposed a new framework of collaborative learning inspired by Optimal Transport theory, where the collaborators aim to increase their local quality based on the information exchanged from other learners. We explained the motivation and the intuition behind our approach and we proposed a new algorithm of collaborative clustering based on the Wasserstein distance. The proposed approach allows to exchange information between collaborators either in vertical or horizontal collaboration. The results are stable and the process stops the collaboration for some learners when their local quality stars to decrease, which prevents common issue of collaborative approaches.

The approach proposed in this paper is the first step into a new family of algorithms for the collaborative leaning task. We plan to develop further collaborative clustering algorithms based on Gromov-Wasserstein distance that ensure the comparison between the distribution coming from heterogeneous spaces, in order to make the collaborative algorithms more flexible, and to improve the quality and the stability of the collaborations.

There are several perspectives to this work. On the short term we are working to improve the approach in order to learn the confidence coefficient at each iteration, according to the diversity and the quality of the collaborators. This could be based on comparisons between sub-sets' distributions using the Wasserstein distance. This would lead us to another extension where the interaction between collaborators could be modeled as graph in a Wasserstein space, which would allow the construction of a theoretical proof of convergence. 

\bibliographystyle{apalike}
\bibliography{references}

\end{document}